\documentclass{article} 
\usepackage{iclr2025_conference,times}
\usepackage{CJKutf8}

\usepackage{amsmath,amsfonts,bm}

\def\eqref#1{equation~\ref{#1}}

\def\1{\bm{1}}

\DeclareMathAlphabet{\mathsfit}{\encodingdefault}{\sfdefault}{m}{sl}
\SetMathAlphabet{\mathsfit}{bold}{\encodingdefault}{\sfdefault}{bx}{n}

\usepackage{hyperref}
\usepackage{url}

\usepackage{pifont}
\usepackage{float}
\usepackage{graphicx}
\usepackage{subfigure}
\usepackage{wrapfig}
\usepackage{amsmath}
\usepackage{amsfonts}
\usepackage{amsthm}

\usepackage{bm}
\usepackage{multirow}
\usepackage{multicol}
\usepackage{tabularx}
\usepackage{booktabs}
\usepackage{setspace}
\usepackage{xcolor,colortbl}
\usepackage{csquotes}
\usepackage{makecell}

\usepackage{fontawesome5}

\definecolor{Gold}{rgb}{1, 0.88, 0.22}
\definecolor{Silver}{rgb}{0.87, 0.87, 0.87}
\definecolor{Bronze}{rgb}{0.88, 0.62, 0.40}

\colorlet{GoldD}{Gold!95!black}
\colorlet{SilverD}{Silver!95!black}
\colorlet{BronzeD}{Bronze!95!black}
\newcommand{\mc}[2]{\textbf{\textcolor{#1D}{#2}}}

\newcommand{\medalbox}[2]{\begingroup\setlength{\fboxsep}{0.5pt}\colorbox{#1}{\strut #2}\endgroup}

\usepackage{tcolorbox}
\tcbuselibrary{skins}

\newtcolorbox{promptbox}[1][]{
  enhanced,
  colback=white,
  colframe=gray!70!black,
  coltitle=white,
  title=#1,
  fonttitle=\bfseries\large,
  boxrule=0.5pt,
  arc=2mm,
  left=1em, right=1em, top=1em, bottom=1em,
  attach boxed title to top left={xshift=0.5em,yshift=-1.2em},
  boxed title style={
    colback=gray!70!black,
    arc=2mm,
    outer arc=0pt,
  },
  fontupper=\rmfamily
}

\definecolor{stepgray}{gray}{0.9}
\definecolor{lightblue}{rgb}{0.53, 0.81, 92}
\definecolor{lightred}{rgb}{1, 0.9, 0.9}
\definecolor{deepblue}{rgb}{0, 0.4470, 0.7410}
\definecolor{deepyellow}{rgb}{0.9290, 0.6940, 0.1250}
\definecolor{deepgreen}{rgb}{0,0.5,0}
\usepackage{enumitem}
\setlist[itemize]{left=1em}

\title{HiPhO: How Far Are (M)LLMs from Humans \\in the Latest High School Physics Olympiad Benchmark?}

\author{
    Fangchen Yu$^{1,2}$\thanks{Equal contribution. $^{ \dagger}$Corresponding author.}~~, 
    Haiyuan Wan$^{1,4 \, *}$,
    Qianjia Cheng$^{1,5 \, *}$, 
    Yuchen Zhang$^{1,6}$, 
    Jiacheng Chen$^{1}$,\\ \textbf{
    Fujun Han$^{2}$,
    Yulun Wu$^{1,5}$,
    Junchi Yao$^{1,7}$,
    Ruilizhen Hu$^{2}$,
    Ning Ding$^{1,4}$,
    Yu Cheng$^{1,3}$,} \\ \textbf{ 
    Tao Chen$^{8}$,
    Lei Bai$^{1}$,
    Dongzhan Zhou$^{1 \dagger}$,
    Yun Luo$^{1 \dagger}$,
    Ganqu Cui$^{1 \dagger}$,
    Peng Ye$^{1,3 \dagger}$}\\
    $^1$Shanghai AI Laboratory,
    $^2$CUHK-Shenzhen,
    $^3$CUHK,
    $^4$Tsinghua University,
    $^5$Zhejiang University,\\
    $^6$Peking University,
    $^7$University of Electronic Science and Technology of China,
    $^8$Fudan University
}

\NewDocumentCommand{\ganqu}
{ mO{} }{\textcolor{blue}{\textsuperscript{\textit{ganqu}}\textsf{\textbf{\small[#1]}}}}

\iclrfinalcopy 
\begin{document}

\maketitle

\begin{abstract}

    Recently, the physical capabilities of (M)LLMs have garnered increasing attention. However, existing benchmarks for physics suffer from two major gaps: they neither provide systematic and up-to-date coverage of real-world physics competitions such as physics Olympiads, nor enable direct performance comparison with humans. To bridge these gaps, we present \textbf{\textsc{HiPhO}}, the first benchmark dedicated to high school physics Olympiads with human-aligned evaluation. Specifically, \textsc{HiPhO} highlights three key innovations. \textbf{(1) Comprehensive Data:} It compiles 13 latest Olympiad exams from 2024–2025, spanning both international and regional competitions, and covering mixed modalities that encompass problems spanning text-only to diagram-based. \textbf{(2) Professional Evaluation:} We adopt official marking schemes to perform fine-grained grading at both the answer and step level, fully aligned with human examiners to ensure high-quality and domain-specific evaluation. \textbf{(3) Comparison with Human Contestants:} We assign gold, silver, and bronze medals to models based on official medal thresholds, thereby enabling direct comparison between (M)LLMs and human contestants. Our large-scale evaluation of 30 state-of-the-art (M)LLMs shows that: across 13 exams, open-source MLLMs mostly remain at or below the bronze level; open-source LLMs show promising progress with multiple golds; closed-source reasoning MLLMs can achieve 6 to 12 gold medals; and most models still have a significant gap from full marks. These results highlight the performance gap between open-source models and top students, the strong reasoning abilities of closed-source models, and the remaining room for improvement. \textsc{HiPhO}, a human-aligned Olympiad benchmark for multimodal physical reasoning, is open-source at \url{https://github.com/SciYu/HiPhO} with a public leaderboard at \url{https://phyarena.github.io/}.

\end{abstract}


\begin{figure}[H]
\vspace{-5mm}
    \centering
    \includegraphics[width=.7\linewidth]{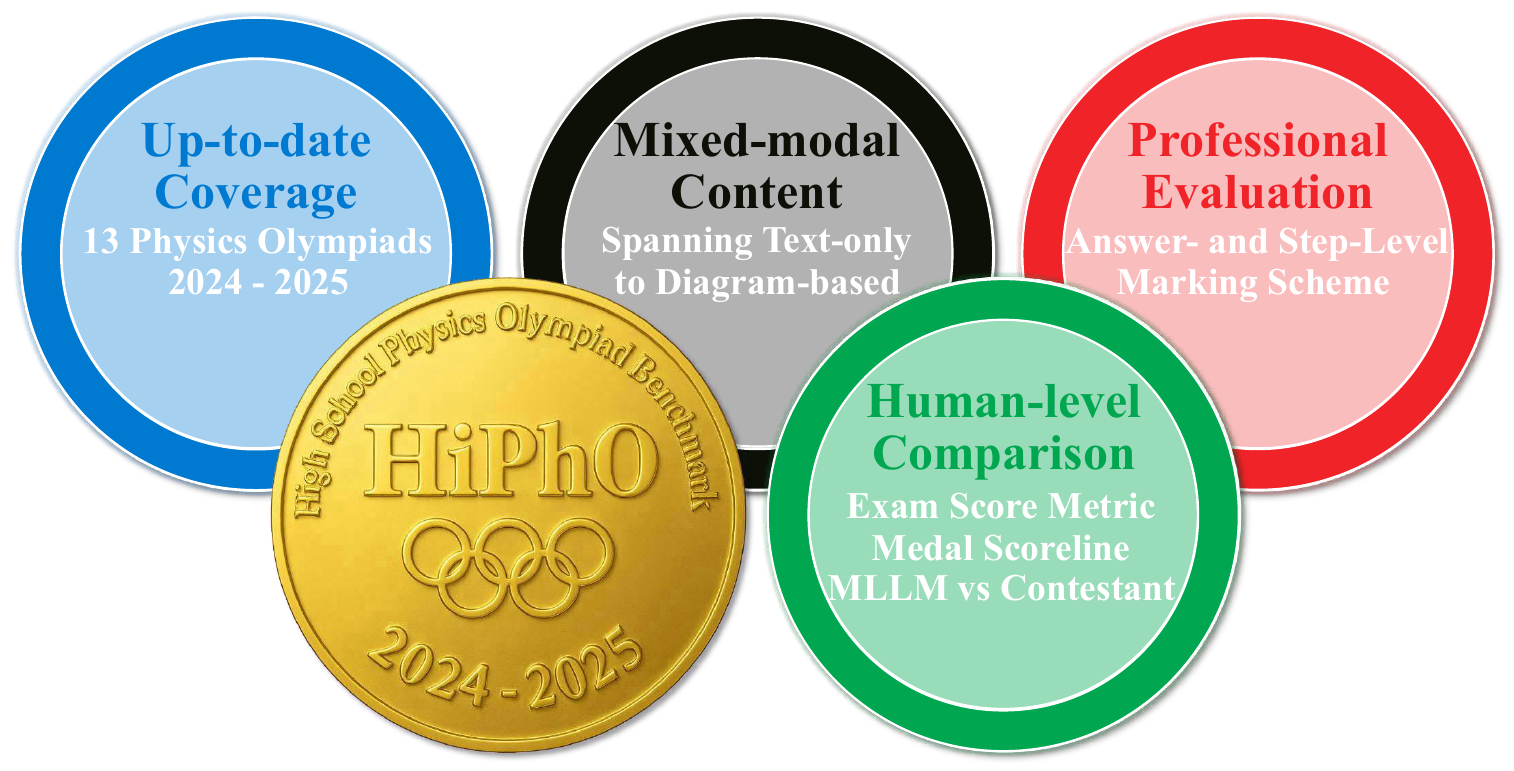}
    \vspace{-2mm}
    
    \caption{The overview of our \textsc{HiPhO} (High School Physics Olympiad) benchmark.}
    \label{fig:overview}
\end{figure}

\begin{figure}
    \centering
    \includegraphics[width=1\linewidth]{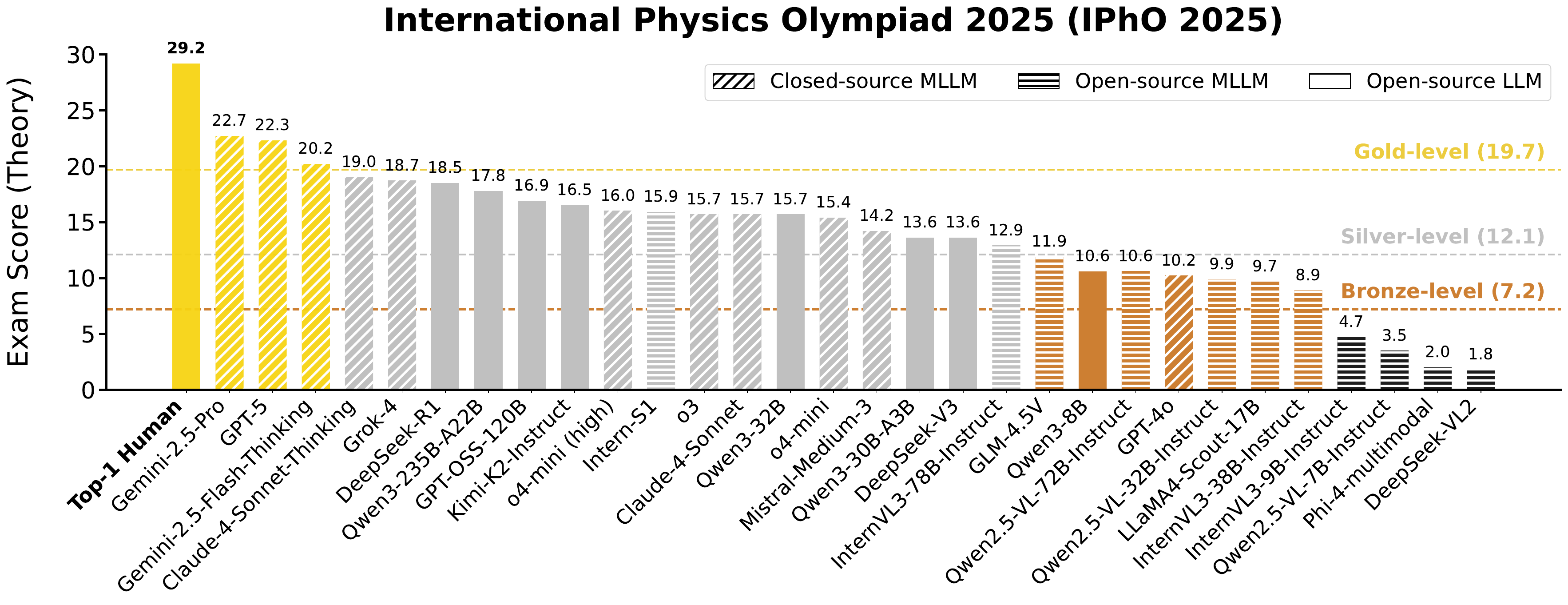}
    \vspace{-7mm}
    
    \caption{Model performance on the International Physics Olympiad 2025 (IPhO 2025)\protect\footnotemark. We show the performance gap between state-of-the-art (M)LLMs and the top-1 human contestant.}
    \label{fig:results_ipho25}
    \vspace{-1mm}
\end{figure}
\footnotetext{In this work, medal thresholds are determined based on the theoretical exam scores of human medalists.}

\newpage
\section{Introduction}

\begin{center}
\textit{\enquote{Physics is essentially an intuitive and concrete science.}} \;\;\;
--- \textsc{Albert Einstein}
\end{center}

Large language models (LLMs) and Multimodal LLMs (MLLMs) have recently attracted increasing attention for their physical reasoning capabilities. However, existing datasets for physics remain limited in scope, lacking both systematic coverage of physics Olympiads and direct comparison with human contestants. By contrast, mathematical Olympiads have already received extensive attention—models such as GPT-5 have achieved over 90\% accuracy on challenging benchmarks like AIME-2025~\citep{2025aime}, and DeepMind's Gemini (``Deep Think'') even reached gold-medal level at the 2025 International Mathematical Olympiad (IMO). As the crown jewel of high school physics, Olympiad problems offer a uniquely rigorous testbed that remains underexplored, underscoring the need for a new benchmark with systematic evaluation and human-aligned comparison.

Physics Olympiads, such as the International Physics Olympiad (IPhO), represent the pinnacle of high school physics competitions. Unlike mathematical Olympiads, they require a deep understanding of real-world physical principles, the derivation of abstract physics formulas, and the ability to reason with complex multimodal diagrams. These characteristics make physics Olympiads an ideal testbed for evaluating whether (M)LLMs can perform authentic visual and physical reasoning. Moreover, they offer the unique advantage of enabling direct comparison with human contestants.

However, current Olympiad-related physics datasets exhibit several critical limitations:
\textbf{(1) Outdated coverage:} Datasets like OlympiadBench \citep{2024olympiadbench} includes IPhO problems only up to 2021 and Asian Physics Olympiad (APhO) problems up to 2015, omitting the most recent two years of exams.
\textbf{(2) Lack of multimodal content:} Physics Olympiad problems often involve complex diagrams. However, datasets such as PHYBench \citep{2025phybench} and PHYSICS \citep{2025physics2} are entirely text-based.
\textbf{(3) Limited evaluation quality:} Most datasets stop at coarse answer-level evaluation, without step-level grading aligned with official marking schemes.
\textbf{(4) No human-level comparison:} Existing datasets typically report accuracy, without exam-score-based comparisons against human contestants in real-world physics competitions.

To address these limitations, we introduce \textbf{\textit{\textsc{HiPhO}}} (High School Physics Olympiad), the first benchmark dedicated to recent physics Olympiads with human-aligned evaluation. As illustrated in Fig.~\ref{fig:overview}, \textsc{HiPhO} incorporates four key improvements: \textbf{(1) Up-to-date coverage:} It compiles the latest 13 Olympiad exams from 2024–2025, including both international and regional contests. \textbf{(2) Mixed-modal content:} Problems are collected as complete exams, spanning text-only to diagram-based, with fine-grained categorization of diagram types (see Fig.~\ref{fig:problem_type}). \textbf{(3) Professional evaluation:} We adopt official marking schemes to perform fine-grained grading at both the answer and step level, fully aligned with human examiners to ensure rigorous and domain-specific assessment. \textbf{(4) Human-level comparison:} We compute full exam scores for models and map them to gold, silver, and bronze medal thresholds, thereby enabling direct comparison with human contestants.

\begin{table}[t]
\centering
\caption{Comparison of physics benchmarks across key dimensions, including Olympiad coverage, modalities, evaluation, metrics, and use of medal scores. \textsc{HiPhO} introduces three innovations: (1) fine-grained modality categorization: 1 = text-only, 2 = text + illustration figure, 3 = text + variable figure, 4 = text + data figure (see Fig.~\ref{fig:problem_type}); (2) step-level grading using official marking schemes; and (3) the first benchmark to compare model performance with medal cutoffs and contestant scores.}
\label{tab:benchmark}
\setlength{\tabcolsep}{2pt}
\small
\resizebox{\textwidth}{!}{%
\begin{tabular}{lcccccc}
\toprule
\multirow{2}{*}{Benchmark} & \multirow{2}{*}{Multimodal}~ & Modality & Evaluation & Performance & Medal \\
 & & Type & Method & Metric & Score \\
\midrule
\multicolumn{6}{c}{\it Non-Olympiad Physics Benchmarks or Subsets} \\
GPQA {\citep{2024gpqa}} & \ding{55} & 1 & Answer & Acc. & \ding{55} \\
HLE {\citep{2025hle}} & \ding{55} & 1 & Answer & Acc. & \ding{55} \\
UGPhysics {\citep{2025ugphysics}} & \ding{55} & 1 & Answer & Acc. & \ding{55} \\
CMPhysBench {\citep{2025cmphysbench}} & \ding{55} & 1 & Answer & Acc., SEED Score & \ding{55} \\
PhyX {\citep{2025phyx}} & \ding{51} & 2 & Answer & Acc. & \ding{55} \\
PhysUniBench {\citep{2025physunibench}} & \ding{51} & 2 & Answer & Acc. & \ding{55} \\
\midrule
\multicolumn{6}{c}{\it Olympiad-related Physics Benchmarks or Subsets} \\
PHYBench {\citep{2025phybench}} & \ding{55} & 1 & Answer & Acc., EED Score & \ding{55} \\
PHYSICS {\citep{2025physics2}} & \ding{55} & 1 & Answer & Acc. & \ding{55} \\
OlympiadBench {\citep{2024olympiadbench}} & \ding{51} & 1,2 & Answer & Acc. & \ding{55} \\
OlympicArena {\citep{2024olympicarena}} & \ding{51} & 1,2 & Answer+{\color{black}Model Step$^1$} & Acc. & \ding{55} \\
PhysReason {\citep{2025physreason}} & \ding{51} & 1,2 & Answer+{\color{black}Error Step$^2$} & Acc. & \ding{55} \\
SeePhys {\citep{2025seephys}} & \ding{51} & 1,2,4 & Answer & Acc. & \ding{55} \\
\midrule
\multicolumn{6}{c}{\textit{\textbf{Olympiad-focused Physics Benchmark}}} \\
\textbf{\textsc{HiPhO} (Ours)} & \ding{51} & 1,2,3,4 & Answer+{\color{black}Marking Step$^3$} & Exam Score & \ding{51} \\
\bottomrule
\end{tabular}%
}

\vspace{0.5em}
\begin{minipage}{\textwidth}
\scriptsize
$^1$GPT4-generated steps; 
$^2$First-error-step detection in model output; 
$^3$Official marking scheme with scoring per predefined point.\\
\end{minipage}
\vspace{-7.2mm}
\end{table}


In our large-scale evaluation of 30 state-of-the-art (M)LLMs, we observe a clear performance hierarchy. \textbf{Closed-source reasoning MLLMs} dominate the medal table, winning 6–12 gold medals in 13 Olympiads. However, even the strongest models, such as Gemini-2.5-Pro and GPT-5, still fall short of the very best human contestants, especially in challenging exams like IPhO and EuPhO. In contrast, \textbf{open-source chat MLLMs} failed to secure any gold medals, with most scoring only at or below the bronze threshold. More encouragingly, several \textbf{open-source reasoning (M)LLMs}, including Intern-S1 and DeepSeek-R1, each achieved 4–8 gold medals, particularly in relatively easier exams such as F=MA. Taken together, these results reveal the strong but still non-parity performance of closed-source models, the evident limitations of open-source chat MLLMs, and the promising trajectory of open-source reasoning (M)LLMs in advancing physics problem solving.

In summary, our main contributions are as follows:

$\bullet$ \textbf{Olympiad-focused Benchmark.}
We present \textsc{HiPhO}, the first benchmark dedicated to high school physics Olympiads, comprising 13 Olympiad exam papers from 2024–2025. All problems are expert-curated, structurally extracted, and manually verified to ensure high quality and consistency.

$\bullet$ \textbf{Human-aligned Scoring.}
We evaluate performance using exam scores rather than the commonly used accuracy, applying both answer- and step-level grading based on official marking schemes. This enables direct, medal-standard comparisons with actual human contestants.

$\bullet$ \textbf{Human-level Comparison.} Compared to human contestants, closed-source reasoning MLLMs reach gold in 6–12 exams, while open-source MLLMs stay mostly at bronze; some open-source LLMs show stronger reasoning with multiple golds, yet all remain far from the very top students.
\vspace{-0.8mm}

\section{Related Work}
\vspace{-2mm}

Among Olympiad-related physics datasets (Table~\ref{tab:benchmark}), \textit{PHYBench} \citep{2025phybench} and \textit{PHYSICS} \citep{2025physics2} are text-only, while \textit{OlympiadBench} \citep{2024olympiadbench} and \textit{OlympicArena} \citep{2024olympicarena} cover outdated exams. Multimodal datasets such as \textit{SeePhys} \citep{2025seephys} and \textit{PhysReason} \citep{2025physreason} remain restricted in scope (e.g., IPhO or CPhO). By contrast, \textsc{HiPhO} compiles 13 recent Olympiads (2024–2025), spanning problems with mixed modalities. More importantly, whereas existing datasets generally stop at coarse answer-level accuracy, \textsc{HiPhO} introduces fine-grained evaluation based on official marking schemes and adopts exam scores as the metric, enabling direct comparison between (M)LLMs and human contestants.
\begin{figure}[t]
    \centering
    \includegraphics[width=1\linewidth]{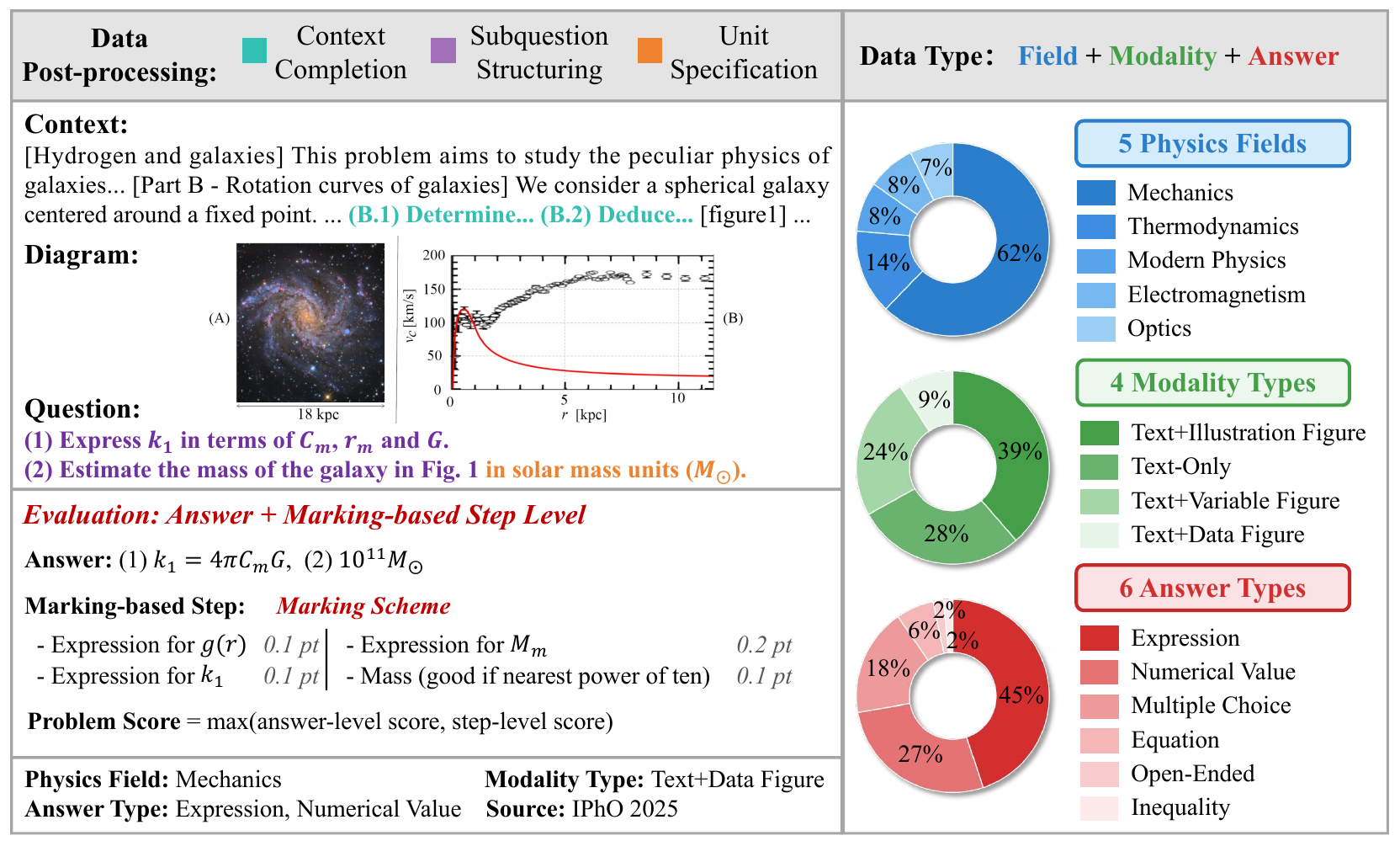}
    
    \caption{Framework and statistics of the \textsc{HiPhO} benchmark. \textbf{Top left:} An example problem with three post-processing steps—context completion, subquestion structuring, and unit specification. \textbf{Bottom left:} Answer + step-level evaluation based on official marking schemes. \textbf{Right:} Dataset composition across five physics fields, four modality types, and six answer types.}
    \label{fig:diagram_statistic}
\end{figure}

\newpage
\section{The \textsc{HiPhO} Benchmark for High School Physics Olympiad}

\subsection{Overview}

\begin{wraptable}{r}{0.35\textwidth}
\centering
\small
\caption{Benchmark Statistics.}
\label{tab:benchmark_statistics}
\setlength{\tabcolsep}{3pt}
\begin{tabular}{lr}
\toprule
\textbf{Statistic} & \textbf{Count} \\
\midrule
\multicolumn{2}{c}{\textit{Data Level}} \\
Total Problems & 360 \\
\quad - Total Subquestions & 519 \\
Total Physics Fields & 5 \\
Total Modality Types & 4 \\
Total Answer Types & 6 \\
Total Difficulty Levels & 3 \\
Language (EN : ZH) & 308:52 \\ 
\midrule
\multicolumn{2}{c}{\textit{Evaluation Level}} \\
Total Exam Papers   & 13 \\
\quad - With Medal Scorelines   & 13 \\
\quad - With Marking Schemes    & 7  \\
\bottomrule
\end{tabular}
\end{wraptable}

We introduce \textsc{HiPhO}, the first high school physics Olympiad benchmark designed to compare the reasoning capabilities of (M)LLMs against human contestants. It contributes along three key dimensions: (1) a comprehensive and up-to-date dataset from real Olympiad problems, (2) professional evaluation using answer- and step-level grading aligned with official marking schemes, and (3) human-level comparison by mapping model scores to official medal thresholds.

\textbf{Dataset Perspective.} As shown in Table~\ref{tab:benchmark_statistics}, \textsc{HiPhO} includes 360 problems and 519 subquestions from 13 Olympiad-level exam papers held in 2024–2025, making it the most up-to-date benchmark. All problems are carefully verified by expert annotators to ensure accuracy. Each problem is categorized along two axes: \textbf{(1) physics taxonomy}—covering five fields (Mechanics, Electromagnetism, Thermodynamics, Optics, and Modern Physics); and \textbf{(2) modality type}—spanning four formats (Text-Only, Text+Illustration Figure, Text+Variable Figure, and Text+Data Figure; see Fig.~\ref{fig:problem_type}), enabling detailed analysis of both physical and visual reasoning.

\textbf{Evaluation Perspective.} \textsc{HiPhO} introduces a new evaluation method that combines answer-level correctness with step-level assessment based on \textit{official marking schemes}—the first benchmark to do so. Beyond fine-grained scoring, we use \textit{exam score} instead of accuracy to provide a more faithful measure of overall exam performance. This enables direct comparison with actual student scores and official medal thresholds, providing the first quantitative analysis of the performance gap between state-of-the-art (M)LLMs and top human contestants in real-world physics competitions.


\newpage
\subsection{Dataset Construction}

\begin{figure}[t]
    \centering
    \includegraphics[width=1\linewidth]{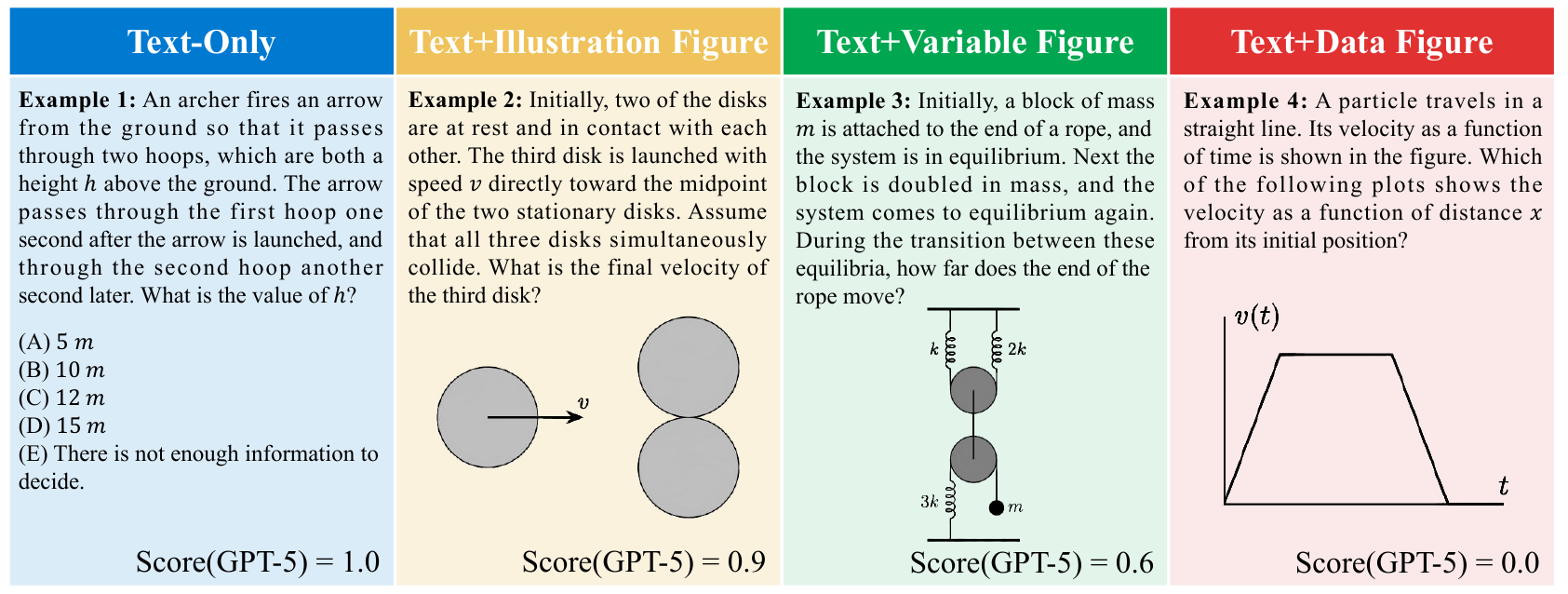}
    \vspace{-4.8mm}
    
    \caption{Examples of the four modality types in the \textsc{HiPhO} benchmark: Text-Only (TO), Text+Illustration Figure (TI), Text+Variable Figure (TV), and Text+Data Figure (TD). TO and TI rely mainly on text, whereas TV and TD require understanding variables or data, making them more challenging for visual reasoning, as reflected in GPT-5's scores (each example worth 1.0 point).}
    \label{fig:problem_type}
\end{figure}

\textbf{Up-to-date Coverage.} \textsc{HiPhO} comprises 13 exam papers from seven major Physics Olympiads (PhOs) and competitions in 2024–2025 (see Appendix~\ref{appendix:data_source}). These were selected based on their influence and the availability of human scores\footnote{Notably, CPhO, USAPhO, and APhO-2024 were excluded due to the absence of human scores.}. Compared with existing datasets, \textsc{HiPhO} offers superior timeliness and breadth: OlympiadBench includes older problems (e.g., APhO-2015) prone to contamination, while PhysReason covers only IPhOs. In contrast, \textsc{HiPhO} integrates the most recent 2024–2025 Olympiad exams. Supported by robust data processing tools, it can be efficiently maintained and expanded with new problems each year, ensuring a comprehensive and continuously updated benchmark. We further categorize the included exams into three difficulty levels:
\begin{itemize}
    \item \textbf{Hard:} \textit{IPhO} (International PhO), \textit{APhO} (Asian PhO), \textit{EuPhO} (European PhO)
    
    \item \textbf{Medium:} \textit{NBPhO} (Nordic-Baltic PhO), \textit{PanPhO} (Pan Pearl River Delta PhO)
    
    \item \textbf{Easy:} \textit{PanMechanics} (Pan Pearl River Delta Mechanics Test), \textit{F=MA}
\end{itemize}

\textbf{High-Quality Extraction Pipeline.} We collect official PDF exam papers from Olympiad websites\footnote{E.g., IPhO 2024–2025: \url{https://ipho.olimpicos.net/}; other websites listed in Appendix~\ref{appendix:data_source}.} and process them through a structured pipeline to ensure high-quality extraction. \textbf{(1) Data Extraction:} PDFs are converted into markdown files using OCR tools, preserving LaTeX formatting for physical expressions. \textbf{(2) QA Matching:} Question indices are used to align each problem with its corresponding answer. \textbf{(3) Human Verification:} Unlike large-scale datasets that rely on automated or sampled checks, \textsc{HiPhO} manually verifies every QA pair. Experts fix OCR errors, correct mismatches, and validate answers to ensure consistency with the original exams. \textbf{(4) Marking Scheme Structuring:} For exams with official marking schemes, we extract step-level criteria and convert them into clear, model-readable rubrics for fine-grained evaluation. \textbf{(5) Post-processing:} Verified QAs are then refined in terms of context, question structure, and unit clarity, as detailed below.

\textbf{Post-Processing and QA Refinement.} To simulate real exam conditions and improve evaluation accuracy, we apply three key post-processing steps, rarely seen in existing datasets. \textbf{(1) Context Completion:} Many Olympiad problems include long stems followed by interdependent subquestions. To ensure contextual coherence, we merge relevant prior content into the context field. \textbf{(2) Subquestion Structuring:} Subquestions are often semantically linked and jointly graded in official marking schemes. We preserve their structure and explicitly label each part (e.g., “Find the speed and acceleration” $\rightarrow$ “(1) Find the speed. (2) Find the acceleration”) to guide ordered model responses and reduce omissions. \textbf{(3) Unit Specification:} We explicitly clarify required units within the question (e.g., “Find the speed in $m/s$”) to avoid penalization from trivial unit mismatches, enhancing scoring fairness and accuracy without altering problem difficulty.

\textbf{Comprehensive Data Annotation.} To support fine-grained analysis, we annotate the dataset across three dimensions. \textbf{(1) Physics Domain Categorization:} Problems are labeled by five major fields—Mechanics, Electromagnetism, Thermodynamics, Optics, and Modern Physics.
\textbf{(2) Visual Modality Classification:} As shown in Fig.~\ref{fig:problem_type}, we categorize problems into four types based on the role of figures, reflecting varying levels of visual reasoning:
\begin{itemize} 
\vspace{-2mm}
    \item \textit{Text-Only (TO)}: Problems are stated entirely in text with no figures.

    \item \textit{Text+Illustration Figure (TI)}: Figures depict scenarios, while the text provides the description.
    
    \item \textit{Text+Variable Figure (TV)}: Figures specify key variables or spatial configurations.  
    
    \item \textit{Text+Data Figure (TD)}: Figures present data, plots, or functions that are not given in the text.
\end{itemize}

\textbf{(3) Answer Type Annotation:} At the subquestion level, answers are classified into six types, including Expression, Numerical Value, Multiple Choice, Equation, Inequality, and Open-Ended. As shown in Fig.~\ref{fig:diagram_statistic}, \textit{expression} is the most common (45\%), e.g., ``express $F$ in terms of $\alpha$ and $\beta$.'' The next most frequent type is \textit{numerical values} (27\%), e.g., ``calculate the value of $F$ in $\mathrm{N}$.'' Together, these two types assess models' ability in both symbolic reasoning and precise calculation.


\section{Evaluation Method}

\textbf{Comprehensive Evaluation Framework.} Our framework integrates both \textit{answer-level} and \textit{step-level} scoring, closely mirroring the grading approach of human examiners. The process is as follows:
\textbf{(1) Answer-level (coarse-grained) scoring:} We first apply rule-based math-verifier \citep{math_verify} to determine whether the extracted model answer matches the ground truth. If the check fails, possibly due to equivalent expressions, we invoke a strong model to assess correctness. 
\textbf{(2) Step-level (fine-grained) scoring:} We compare the model's solution with the official marking scheme. A strong model awards partial credit for correctly completed steps.
The final score for each problem is
\[
\text{Problem Score} = \max(\text{answer-level score}, \text{step-level score}),
\]
following human grading conventions: a correct final answer receives full credit while an incorrect one is still graded fairly through intermediate steps. For problems with multiple official solutions (e.g., NBPhO), we score all marking schemes and take the highest result to ensure fairness across alternative solving paths. The exam score is then obtained by summing the scores of all problems.

\textbf{Key Improvements in Evaluation.} Our evaluation introduces three major enhancements. 
\textbf{(1) Step-level scoring:} Most datasets assess only the final answer (Table~\ref{tab:benchmark}), awarding zero credit if it is incorrect. We incorporate step-level scoring to recognize partial correctness.
\textbf{(2) Marking-scheme alignment:} Unlike prior step-level methods (e.g., OlympicArena generates steps with GPT-4), we are the \textbf{first} to extract steps strictly from the official marking scheme, awarding points based on explicit marking criteria for greater rigor and consistency.  
\textbf{(3) Stronger grader:} We employ Gemini-2.5-Flash as the grading model. Previous works \citep{2025ugphysics,2025physics1} often rely on GPT-4o, whose competition scores are relatively low, making it less suitable for evaluating Olympiad-level problems. Compared to GPT-4o, Gemini-2.5-Flash provides step-level scores more closely aligned with human experts, with under 1-point differences in the examples (see Appendix~\ref{appendix:model_score_human}).

\begin{table}[t]
\centering
\caption{Evaluation results on the \textsc{HiPhO} benchmark (13 physics Olympiads from 2024--2025) using the \textit{exam score} metric. \medalbox{Gold!50}{Gold}, \medalbox{Silver!70}{Silver} and \medalbox{Bronze!40}{Bronze} indicate scores above the respective thresholds. Models are ranked by medal counts; \textbf{bold} is the highest score, and \underline{underline} is the second highest. Here, only the theoretical parts of exams are used, hence Full Mark (Model) $\leq$ Full Mark (Human).}
\label{tab:results}
\small
\resizebox{\textwidth}{!}{%
\setlength{\tabcolsep}{2.3pt}
\begin{tabular}{lccccccccccccc|ccc}
\toprule
\textbf{Physics Olympiad} & \multicolumn{2}{c}{\textbf{IPhO}} & \textbf{APhO} & \multicolumn{2}{c}{\textbf{EuPhO}} & \multicolumn{2}{c}{\textbf{NBPhO}} & \multicolumn{6}{c|}{\textbf{PanPhO}~~
\textbf{\footnotesize PanMechanics} ~~ \textbf{F=MA}} & \multicolumn{3}{c}{\textbf{Medal}} \\
\textbf{Year} & 2025 & 2024 & 2025 & 2025 & 2024 & 2025 & 2024 & 2025 & 2024 & 2025 & 2024 & 2025 & 2024 & \multicolumn{3}{c}{\textbf{Table}} \\ \midrule
Full Mark (Human) & 30.0 & 30.0 & 30.0 & 30.0 & 30.0 & 72.0 & 72.0 & 100.0 & 100.0 & 100.0 & 100.0 & 25.0 & 25.0 \\
Full Mark (Model) & 29.4 & 29.3 & 30.0 & 29.0 & 28.0 & 43.5 & 50.0 & 100.0 & 98.0 & 100.0 & 100.0 & 25.0 & 25.0 \\
Top-1 Score (Human) & 29.2 & 29.4 & 30.0 & 27.0 & 30.0 & 53.2 & 40.8 & 81.0 & 66.5 & 62.0 & 51.0 & 25.0 & 24.0 \\
Top-1 Score (Model) & 22.7 & 25.9 & 27.9 & 14.9 & 21.9 & 34.1 & 35.9 & 60.3 & 75.4 & 72.1 & 79.0 & 22.8 & 22.4 \\
\rowcolor{Gold!40}
Gold Medal   & 19.7 & 20.8 & 23.3 & 16.5 & 20.4 & 28.6 & 26.5 & 41.5 & 52.0 & 52.0 & 51.0 & 15.0 & 14.0 & \multicolumn{3}{>{\cellcolor{white}}l}{\textcolor{Gold}{\faMedal}}  \\
\rowcolor{Silver!60}
Silver Medal & 12.1 & 11.1 & 18.7 & 9.8 & 14.2 & 20.1 & 19.4 & 28.5 & 37.5 & 36.0 & 26.0 & 11.0 & 12.0 & \multicolumn{3}{>{\cellcolor{white}}c}{\textcolor{Silver}{\faMedal}}  \\
\rowcolor{Bronze!40}
Bronze Medal & 7.2  & 3.6  & 13.1 & 5.8  & 8.9  & 15.2  & 13.5  & 14.5 & 16.0 & 20.0 & 12.0 & 9.0 & 10.0 & \multicolumn{3}{>{\cellcolor{white}}r}{\textcolor{Bronze}{\faMedal}}  \\ \midrule
\multicolumn{14}{c|}{\textbf{\it Closed-Source Reasoning MLLMs}}\\
Gemini-2.5-Pro & \cellcolor{Gold!35}\textbf{22.7} & \cellcolor{Gold!35}\textbf{25.9} & \cellcolor{Gold!35}\textbf{27.9} & \cellcolor{Silver!60}\textbf{14.9} & \cellcolor{Gold!35}\underline{21.8} & \cellcolor{Gold!35}32.3 & \cellcolor{Gold!35}\textbf{35.9} & \cellcolor{Gold!35}\textbf{60.3} & \cellcolor{Gold!35}64.1 & \cellcolor{Gold!35}69.5 & \cellcolor{Gold!35}70.2 & \cellcolor{Gold!35}\textbf{22.8} & \cellcolor{Gold!35}\underline{22.0} & ~\mc{Gold}{12} & \mc{Silver}{1} & \mc{Bronze}{0}\\
Gemini-2.5-Flash-Thinking & \cellcolor{Gold!35}20.2 & \cellcolor{Gold!35}\underline{23.9} & \cellcolor{Gold!35}\underline{27.4} & \cellcolor{Silver!60}\underline{13.2} & \cellcolor{Gold!35}\textbf{21.9} & \cellcolor{Gold!35}29.0 & \cellcolor{Gold!35}29.3 & \cellcolor{Gold!35}44.6 & \cellcolor{Gold!35}54.9 & \cellcolor{Gold!35}60.5 & \cellcolor{Gold!35}55.9 & \cellcolor{Gold!35}17.8 & \cellcolor{Gold!35}19.1 & ~\mc{Gold}{12} & \mc{Silver}{1} & \mc{Bronze}{0}\\
GPT-5 & \cellcolor{Gold!35}\underline{22.3} & \cellcolor{Silver!60}20.2 & \cellcolor{Gold!35}27.0 & \cellcolor{Silver!60}10.3 & \cellcolor{Gold!35}21.7 & \cellcolor{Gold!35}\underline{32.9} & \cellcolor{Gold!35}32.8 & \cellcolor{Gold!35}\underline{55.9} & \cellcolor{Gold!35}\underline{69.8} & \cellcolor{Gold!35}69.4 & \cellcolor{Gold!35}\textbf{79.0} & \cellcolor{Gold!35}\underline{22.4} & \cellcolor{Gold!35}\textbf{22.4} & ~\mc{Gold}{11} & \mc{Silver}{2} & \mc{Bronze}{0} \\
o3 & \cellcolor{Silver!60}15.7 & \cellcolor{Gold!35}23.7 & \cellcolor{Gold!35}25.9 & \cellcolor{Silver!60}11.4 & \cellcolor{Gold!35}21.6 & \cellcolor{Gold!35}\textbf{34.1} & \cellcolor{Gold!35}\underline{33.5} & \cellcolor{Gold!35}47.3 & \cellcolor{Gold!35}55.9 & \cellcolor{Gold!35}\underline{71.4} & \cellcolor{Gold!35}75.6 & \cellcolor{Gold!35}22.0 & \cellcolor{Gold!35}20.6 & ~\mc{Gold}{11} & \mc{Silver}{2} & \mc{Bronze}{0}\\
Grok-4 & \cellcolor{Silver!60}18.7 & \cellcolor{Gold!35}23.5 & \cellcolor{Gold!35}25.0 & \cellcolor{Silver!60}11.5 & \cellcolor{Gold!35}20.5 & \cellcolor{Silver!60}25.8 & \cellcolor{Gold!35}29.3 & \cellcolor{Gold!35}45.0 & \cellcolor{Gold!35}\textbf{75.4} & \cellcolor{Gold!35}\textbf{72.1} & \cellcolor{Gold!35}\underline{78.6} & \cellcolor{Gold!35}19.8 & \cellcolor{Gold!35}19.8 & ~\mc{Gold}{10} & \mc{Silver}{3} & \mc{Bronze}{0}\\
Claude-4-Sonnet-Thinking & \cellcolor{Silver!60}19.0 & \cellcolor{Gold!35}22.0 & \cellcolor{Gold!35}24.8 & \cellcolor{Bronze!40}9.7 & \cellcolor{Gold!35}20.5 & \cellcolor{Silver!60}28.1 & \cellcolor{Silver!60}25.6 & \cellcolor{Gold!35}43.1 & \cellcolor{Silver!60}39.3 & \cellcolor{Gold!35}57.4 & \cellcolor{Gold!35}61.8 & \cellcolor{Gold!35}19.2 & \cellcolor{Gold!35}20.1 & ~\mc{Gold}{8} & \mc{Silver}{4} & \mc{Bronze}{1}\\
o4-mini & \cellcolor{Silver!60}15.4 & \cellcolor{Gold!35}22.9 & \cellcolor{Silver!60}22.8 & \cellcolor{Silver!60}10.1 & \cellcolor{Gold!35}20.9 & \cellcolor{Silver!60}26.9 & \cellcolor{Gold!35}27.3 & \cellcolor{Silver!60}39.4 & \cellcolor{Silver!60}47.1 & \cellcolor{Gold!35}64.2 & \cellcolor{Gold!35}62.5 & \cellcolor{Gold!35}18.6 & \cellcolor{Gold!35}18.5 & ~\mc{Gold}{7} & \mc{Silver}{6} & \mc{Bronze}{0} \\
o4-mini (high) & \cellcolor{Silver!60}16.0 & \cellcolor{Gold!35}23.7 & \cellcolor{Silver!60}22.9 & \cellcolor{Silver!60}12.0 & \cellcolor{Silver!60}20.1 & \cellcolor{Silver!60}27.4 & \cellcolor{Gold!35}29.8 & \cellcolor{Silver!60}41.4 & \cellcolor{Silver!60}50.9 & \cellcolor{Gold!35}69.1 & \cellcolor{Gold!35}67.3 & \cellcolor{Gold!35}18.6 & \cellcolor{Gold!35}18.8 & ~\mc{Gold}{6} & \mc{Silver}{7} & \mc{Bronze}{0}\\
\multicolumn{14}{c|}{\textbf{\it Open-Source Reasoning MLLMs}}\\
Intern-S1 & \cellcolor{Silver!60}15.9 & \cellcolor{Silver!60}14.2 & \cellcolor{Silver!60}21.7 & \cellcolor{Bronze!40}9.0 & \cellcolor{Silver!60}16.6 & \cellcolor{Silver!60}23.0 & \cellcolor{Silver!60}20.5 & \cellcolor{Silver!60}41.1 & \cellcolor{Silver!60}50.3 & \cellcolor{Gold!35}60.4 & \cellcolor{Gold!35}57.4 & \cellcolor{Gold!35}18.4 & \cellcolor{Gold!35}19.5 & ~\mc{Gold}{4} & \mc{Silver}{8} & \mc{Bronze}{1} \\
GLM-4.5V & \cellcolor{Bronze!40}11.9 & \cellcolor{Bronze!40}4.4 & \cellcolor{Bronze!40}16.2 & \cellcolor{Bronze!40}8.7 & \cellcolor{Bronze!40}14.1 & \cellcolor{Bronze!40}19.5 & \cellcolor{Bronze!40}14.0 & \cellcolor{Bronze!40}18.5 & \cellcolor{Bronze!40}16.0 & \cellcolor{Silver!60}47.8 & \cellcolor{Silver!60}39.0 & \cellcolor{Silver!60}13.0 & \cellcolor{Silver!60}13.8 & ~\mc{Gold}{0} & \mc{Silver}{4} & \mc{Bronze}{9} \\
\arrayrulecolor[gray]{0.7}\midrule\arrayrulecolor{black}
\multicolumn{14}{c|}{\textbf{\it Closed-Source Chat MLLMs}}\\
Claude-4-Sonnet & \cellcolor{Silver!60}15.7 & \cellcolor{Silver!60}19.2 & \cellcolor{Silver!60}22.8 & \cellcolor{Bronze!40}9.5 & \cellcolor{Silver!60}16.5 & \cellcolor{Silver!60}27.5 & \cellcolor{Silver!60}21.3 & \cellcolor{Silver!60}40.4 & \cellcolor{Silver!60}43.3 & \cellcolor{Silver!60}46.5 & \cellcolor{Silver!60}48.5 & \cellcolor{Gold!35}16.8 & \cellcolor{Gold!35}16.5 & ~\mc{Gold}{2} & \mc{Silver}{10} & \mc{Bronze}{1}\\
Mistral-Medium-3 & \cellcolor{Silver!60}14.2 & \cellcolor{Silver!60}14.1 & \cellcolor{Silver!60}19.9 & \cellcolor{Bronze!40}8.5 & \cellcolor{Bronze!40}12.2 & \cellcolor{Silver!60}20.4 & \cellcolor{Silver!60}19.6 & \cellcolor{Silver!60}30.8 & \cellcolor{Bronze!40}28.6 & \cellcolor{Bronze!40}32.9 & \cellcolor{Silver!60}36.1 & \cellcolor{Silver!60}13.9 & \cellcolor{Gold!35}14.1 & ~\mc{Gold}{1} & \mc{Silver}{8} & \mc{Bronze}{4}\\
GPT-4o & \cellcolor{Bronze!40}10.2 & \cellcolor{Bronze!40}9.4 & \cellcolor{Bronze!40}15.1 & \cellcolor{Bronze!40}6.8 & \cellcolor{Bronze!40}9.2 & \cellcolor{Bronze!40}16.4 & 11.7 & \cellcolor{Bronze!40}27.8 & \cellcolor{Bronze!40}22.8 & \cellcolor{Bronze!40}28.2 & \cellcolor{Silver!60}26.5 & \cellcolor{Gold!35}15.0 & \cellcolor{Bronze!40}10.9 & ~\mc{Gold}{1} & \mc{Silver}{1} & \mc{Bronze}{10}\\
\multicolumn{14}{c|}{\textbf{\it Open-Source Chat MLLMs}}\\
InternVL3-78B-Instruct & \cellcolor{Silver!60}12.9 & \cellcolor{Silver!60}12.5 & \cellcolor{Bronze!40}17.7 & \cellcolor{Bronze!40}7.5 & \cellcolor{Silver!60}15.2 & \cellcolor{Silver!60}22.3 & \cellcolor{Silver!60}22.5 & \cellcolor{Bronze!40}26.2 & \cellcolor{Bronze!40}27.4 & \cellcolor{Bronze!40}21.1 & \cellcolor{Silver!60}27.1 & \cellcolor{Silver!60}12.0 & \cellcolor{Silver!60}13.0 & ~\mc{Gold}{0} & \mc{Silver}{8} & \mc{Bronze}{5}\\
Qwen2.5-VL-72B-Instruct & \cellcolor{Bronze!40}10.6 & \cellcolor{Bronze!40}7.2 & \cellcolor{Bronze!40}13.6 & \cellcolor{Bronze!40}6.1 & 8.1 & 13.3 & 11.4 & \cellcolor{Bronze!40}26.8 & \cellcolor{Bronze!40}18.2 & \cellcolor{Bronze!40}24.0 & \cellcolor{Silver!60}28.5 & \cellcolor{Silver!60}13.5 & 9.8 & ~\mc{Gold}{0} & \mc{Silver}{2} & \mc{Bronze}{7} \\
LLaMA4-Scout-17B & \cellcolor{Bronze!40}9.7 & \cellcolor{Bronze!40}9.5 & \cellcolor{Bronze!40}13.1 & 5.4 & \cellcolor{Bronze!40}10.4 & \cellcolor{Silver!60}22.8 & 12.8 & \cellcolor{Bronze!40}26.6 & \cellcolor{Bronze!40}24.1 & \cellcolor{Bronze!40}35.4 & \cellcolor{Silver!60}34.5 & 8.1 & 6.4 & ~\mc{Gold}{0} & \mc{Silver}{2} & \mc{Bronze}{7}\\
Qwen2.5-VL-32B-Instruct & \cellcolor{Bronze!40}9.9 & \cellcolor{Bronze!40}8.2 & \cellcolor{Bronze!40}16.5 & \cellcolor{Bronze!40}6.9 & \cellcolor{Bronze!40}10.0 & \cellcolor{Bronze!40}15.3 & \cellcolor{Bronze!40}14.4 & \cellcolor{Bronze!40}22.5 & \cellcolor{Bronze!40}22.4 & \cellcolor{Bronze!40}28.1 & \cellcolor{Silver!60}29.9 & 7.6 & 4.6 & ~\mc{Gold}{0} & \mc{Silver}{1} & \mc{Bronze}{10} \\
InternVL3-38B-Instruct & \cellcolor{Bronze!40}8.9 & \cellcolor{Bronze!40}7.8 & 12.3 & \cellcolor{Bronze!40}6.1 & 8.3 & 14.0 & 10.6 & \cellcolor{Bronze!40}24.1 & \cellcolor{Bronze!40}20.4 & \cellcolor{Bronze!40}27.5 & \cellcolor{Bronze!40}24.8 & 8.2 & 6.8 & ~\mc{Gold}{0} & \mc{Silver}{0} & \mc{Bronze}{7}\\
InternVL3-9B-Instruct & 4.7 & \cellcolor{Bronze!40}3.7 & 7.2 & 4.2 & 4.1 & 9.4 & 6.0 & 12.4 & 11.0 & 11.6 & \cellcolor{Bronze!40}16.3 & 6.4 & 6.2 & ~\mc{Gold}{0} & \mc{Silver}{0} & \mc{Bronze}{2}\\
Qwen2.5-VL-7B-Instruct & 3.5 & 2.5 & 5.7 & 4.4 & 3.6 & 7.3 & 5.5 & \cellcolor{Bronze!40}14.7 & 7.6 & 9.8 & 11.5 & 4.4 & 3.5 & ~\mc{Gold}{0} & \mc{Silver}{0} & \mc{Bronze}{1}\\
Phi-4-multimodal & 2.0 & 1.6 & 4.2 & 3.6 & 3.6 & 5.0 & 4.5 & 8.3 & 9.0 & 10.0 & 10.1 & 4.4 & 5.0 & ~\mc{Gold}{0} & \mc{Silver}{0} & \mc{Bronze}{0}\\
DeepSeek-VL2 & 1.8 & 0.5 & 2.5 & 3.4 & 3.4 & 5.0 & 3.2 & 5.6 & 4.8 & 7.3 & 6.4 & 5.0 & 3.9 & ~\mc{Gold}{0} & \mc{Silver}{0} & \mc{Bronze}{0}\\
\arrayrulecolor[gray]{0.7}\midrule\arrayrulecolor{black}
\multicolumn{14}{c|}{\textbf{\it Open-Source LLMs}}\\
DeepSeek-R1 & \cellcolor{Silver!60}18.5 & \cellcolor{Gold!35}24.6 & \cellcolor{Gold!35}25.4 & \cellcolor{Silver!60}10.8 & \cellcolor{Gold!35}21.4 & \cellcolor{Silver!60}26.3 & \cellcolor{Silver!60}20.5 & \cellcolor{Gold!35}42.2 & \cellcolor{Silver!60}47.4 & \cellcolor{Gold!35}65.4 & \cellcolor{Gold!35}72.5 & \cellcolor{Gold!35}18.3 & \cellcolor{Gold!35}18.5 & ~\mc{Gold}{8} & \mc{Silver}{5} & \mc{Bronze}{0} \\
Qwen3-235B-A22B & \cellcolor{Silver!60}17.8 & \cellcolor{Gold!35}23.8 & \cellcolor{Gold!35}26.0 & \cellcolor{Bronze!40}9.2 & \cellcolor{Gold!35}21.5 & \cellcolor{Silver!60}28.4 & \cellcolor{Gold!35}31.1 & \cellcolor{Gold!35}42.3 & \cellcolor{Silver!60}49.1 & \cellcolor{Gold!35}63.1 & \cellcolor{Silver!60}44.6 & \cellcolor{Gold!35}18.4 & \cellcolor{Gold!35}17.4 & ~\mc{Gold}{8} & \mc{Silver}{4} & \mc{Bronze}{1} \\
Qwen3-32B & \cellcolor{Silver!60}15.7 & \cellcolor{Silver!60}19.3 & \cellcolor{Gold!35}23.9 & \cellcolor{Silver!60}9.8 & \cellcolor{Gold!35}21.2 & \cellcolor{Gold!35}28.9 & \cellcolor{Silver!60}24.1 & \cellcolor{Silver!60}36.6 & \cellcolor{Silver!60}41.8 & \cellcolor{Gold!35}67.0 & \cellcolor{Gold!35}59.2 & \cellcolor{Gold!35}18.9 & \cellcolor{Gold!35}16.6 & ~\mc{Gold}{7} & \mc{Silver}{6} & \mc{Bronze}{0} \\
Kimi-K2-Instruct & \cellcolor{Silver!60}16.5 & \cellcolor{Silver!60}19.8 & \cellcolor{Gold!35}24.2 & \cellcolor{Silver!60}11.0 & \cellcolor{Silver!60}16.9 & \cellcolor{Silver!60}26.5 & \cellcolor{Silver!60}26.2 & \cellcolor{Silver!60}35.9 & \cellcolor{Silver!60}41.8 & \cellcolor{Gold!35}65.9 & \cellcolor{Gold!35}58.9 & \cellcolor{Gold!35}16.0 & \cellcolor{Gold!35}18.2 & ~\mc{Gold}{5} & \mc{Silver}{8} & \mc{Bronze}{0} \\
GPT-OSS-120B & \cellcolor{Silver!60}16.9 & \cellcolor{Gold!35}21.4 & \cellcolor{Silver!60}22.8 & \cellcolor{Bronze!40}9.1 & \cellcolor{Silver!60}19.9 & \cellcolor{Silver!60}26.0 & \cellcolor{Silver!60}25.8 & \cellcolor{Silver!60}37.4 & \cellcolor{Silver!60}41.8 & \cellcolor{Gold!35}57.1 & \cellcolor{Gold!35}59.7 & \cellcolor{Gold!35}17.8 & \cellcolor{Gold!35}17.6 & ~\mc{Gold}{5} & \mc{Silver}{7} & \mc{Bronze}{1} \\
Qwen3-30B-A3B & \cellcolor{Silver!60}13.6 & \cellcolor{Silver!60}15.4 & \cellcolor{Silver!60}22.7 & \cellcolor{Silver!60}9.8 & \cellcolor{Silver!60}16.5 & \cellcolor{Silver!60}24.7 & \cellcolor{Silver!60}21.7 & \cellcolor{Silver!60}31.9 & \cellcolor{Silver!60}39.5 & \cellcolor{Silver!60}49.9 & \cellcolor{Silver!60}45.0 & \cellcolor{Gold!35}15.5 & \cellcolor{Gold!35}15.0 & ~\mc{Gold}{2} & \mc{Silver}{11} & \mc{Bronze}{0} \\
DeepSeek-V3 & \cellcolor{Silver!60}13.6 & \cellcolor{Silver!60}16.4 & \cellcolor{Silver!60}22.1 & \cellcolor{Bronze!40}7.1 & \cellcolor{Silver!60}17.2 & \cellcolor{Silver!60}21.1 & \cellcolor{Bronze!40}17.3 & \cellcolor{Silver!60}37.2 & \cellcolor{Bronze!40}35.0 & \cellcolor{Silver!60}48.4 & \cellcolor{Silver!60}46.5 & \cellcolor{Silver!60}14.1 & \cellcolor{Gold!35}15.6 & ~\mc{Gold}{1} & \mc{Silver}{9} & \mc{Bronze}{3} \\
Qwen3-8B & \cellcolor{Bronze!40}10.6 & \cellcolor{Silver!60}12.7 & 11.5 & \cellcolor{Bronze!40}7.1 & \cellcolor{Bronze!40}11.9 & \cellcolor{Silver!60}20.1 & \cellcolor{Bronze!40}17.3 & \cellcolor{Bronze!40}26.3 & \cellcolor{Bronze!40}22.3 & \cellcolor{Bronze!40}21.8 & \cellcolor{Bronze!40}22.8 & \cellcolor{Bronze!40}10.8 & \cellcolor{Bronze!40}10.0 & ~\mc{Gold}{0} & \mc{Silver}{2} & \mc{Bronze}{10} \\
\bottomrule
\end{tabular}%
}
\vspace{-4mm}
\end{table}

\section{Experiments}

To systematically evaluate the performance of state-of-the-art (M)LLMs against top-performing students in physics Olympiads, we selected 30 representative models. The experimental setup is detailed in Appendix~\ref{appendix:experimental_setup}, and the medal scorelines are illustrated in Appendix~\ref{appendix:medal_scoreline}.

\textbf{11 Closed-source MLLMs:} GPT-5 \citep{GPT-5}, o3 \citep{o3_o4-mini}, o4-mini (high) \citep{o3_o4-mini}, o4-mini \citep{o3_o4-mini}, GPT-4o \citep{GPT-4o}, Gemini-2.5 Series \citep{Gemini-2.5}, Grok-4 \citep{Grok-4}, Claude-4-Sonnet(-Thinking) \citep{Claude-3.7-Sonnet}, Mistral-Medium-3 \citep{Mistral-Medium-3}.

\textbf{11 Open-source MLLMs:} Intern-S1 \citep{Intern-S1}, InternVL3 Series \citep{InternVL3}, Qwen2.5-VL Series \citep{Qwen2.5-VL}, GLM-4.5V \citep{GLM-4.5V}, 
DeepSeek-VL2 \citep{DeepSeek-VL2}, LLaMA4-Scout-17B \citep{LLaMA4-Scout}, Phi-4-multimodal \citep{Phi-4-multimodal}.

\textbf{8 Open-source LLMs:} GPT-OSS \citep{2025GPT-OSS}, Kimi-K2-Instruct \citep{2025Kimi-K2}, DeepSeek-R1 \citep{2025DeepSeek-R1}, DeepSeek-V3 \citep{2024DeepSeek-V3}, Qwen3 Series \citep{2025Qwen3}.


\newpage
\subsection{Main Results}
\vspace{-0.3mm}

\textbf{What are the most powerful (M)LLMs in physics Olympiads?}  
The medal table in Table~\ref{tab:results} ranks models just like an Olympiad. The top five positions are occupied by \textbf{Gemini-2.5-Pro}, \textbf{Gemini-2.5-Flash-Thinking}, \textbf{GPT-5}, \textbf{o3}, and \textbf{Grok-4}. However, in the challenging IPhO-2025, only three of them reached the gold threshold, with the top two separated by merely 0.4 points. This narrow margin highlights the intense competition at the frontier of closed-source model performance.

\textbf{How far are (M)LLMs from top-performing students?} \textsc{HiPhO} enables human-level comparisons. \textbf{(1) Closed-source reasoning MLLMs} collected 6–12 gold medals across the 13 exams, yet still fell short of the very best students—for instance, in IPhO-2025 the top human scored \textbf{29.2/30}, while the best model achieved only \textbf{22.7/29.4}. \textbf{(2) Open-source MLLMs} mostly remained at or below the bronze threshold, with Intern-S1 the only exception to reach gold. \textbf{(3) Open-source LLMs} generally outperformed open-source MLLMs, earning gold in easier contests such as F=MA and even reaching the gold threshold in IPhO-2024, but still lagged well behind the very top students.

\textbf{Can open-source models catch up with closed-source champions?}  
While closed-source reasoning models still dominate, recent progress in the open-source community is noteworthy. Intern-S1 distinguished itself as the only open-source MLLM to win gold, obtaining four medals. Even more impressively, open-source LLMs including DeepSeek-R1 and Qwen3-235B-A22B both achieved eight gold medals. These advances reveal both the enduring gap with closed-source leaders and the emerging potential of open-source research to narrow it.

\begin{figure}[t]
    \centering
    \includegraphics[clip, trim=30 0 30 0, width=1\linewidth]{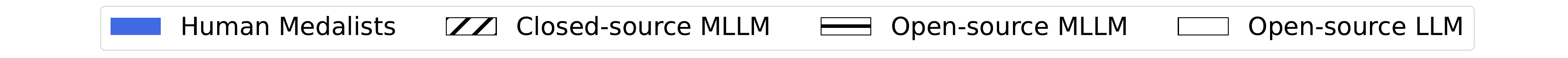}
    \includegraphics[width=0.328\linewidth]{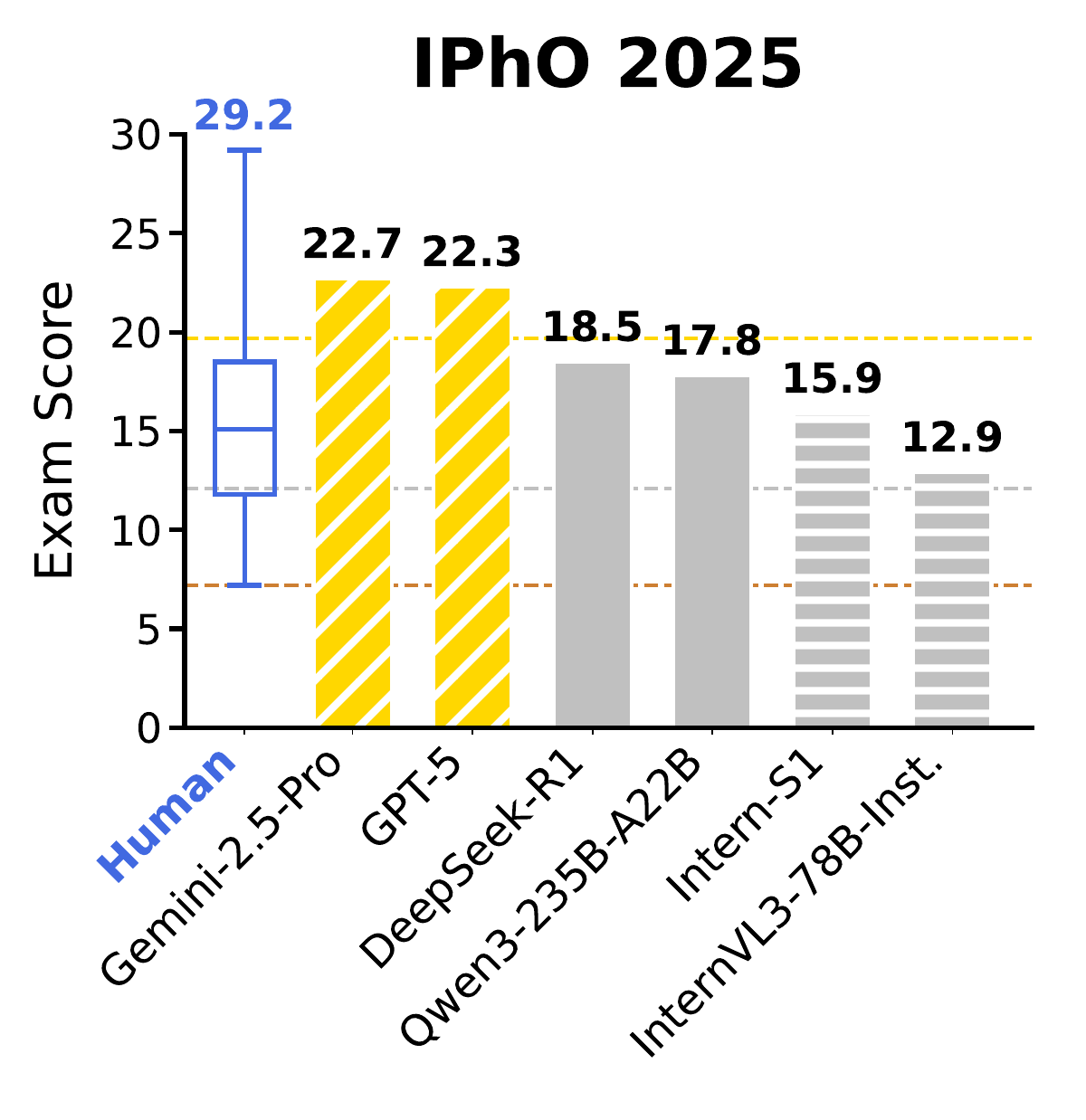}
    \includegraphics[width=0.328\linewidth]{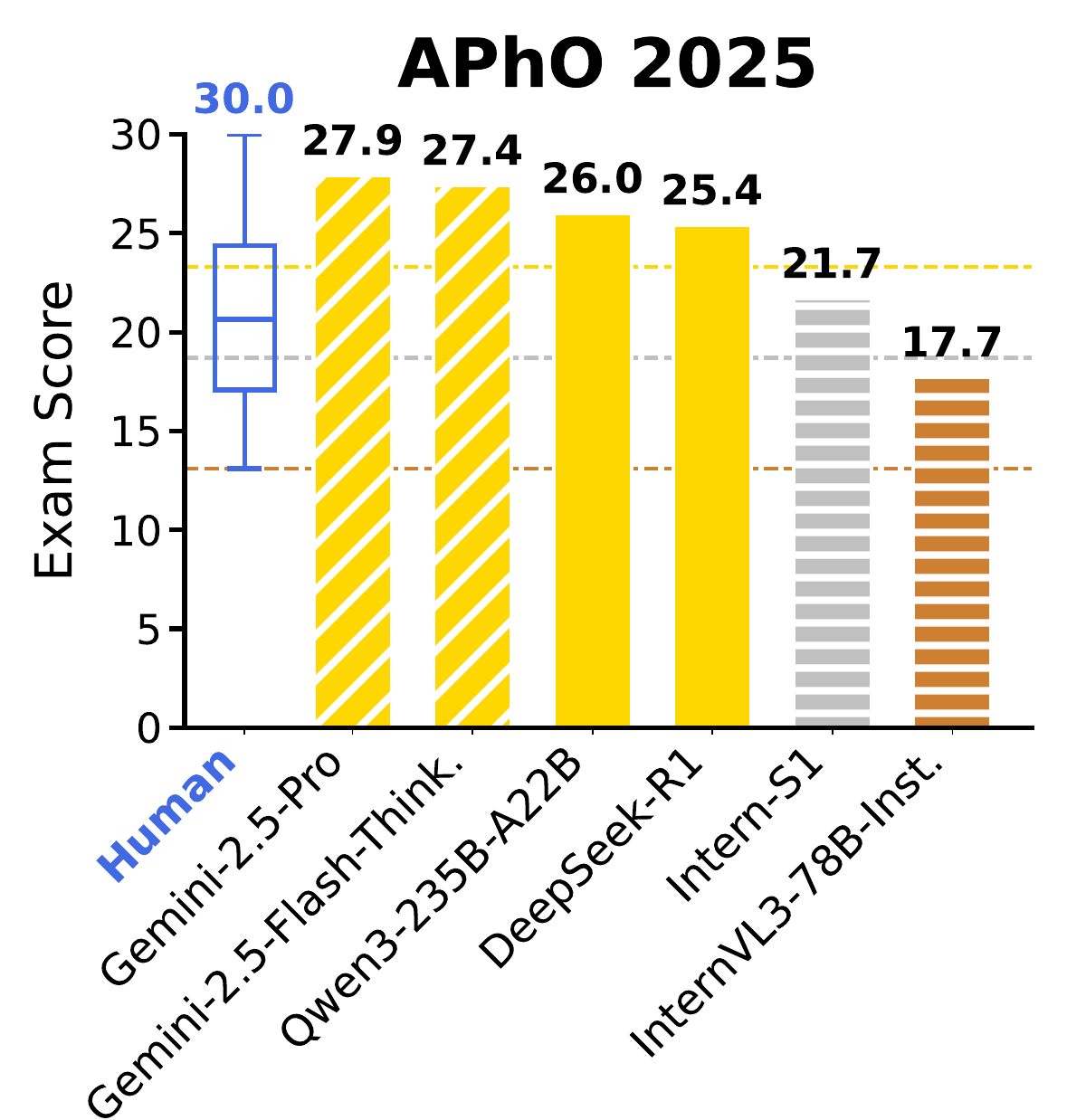}
    \includegraphics[width=0.328\linewidth]{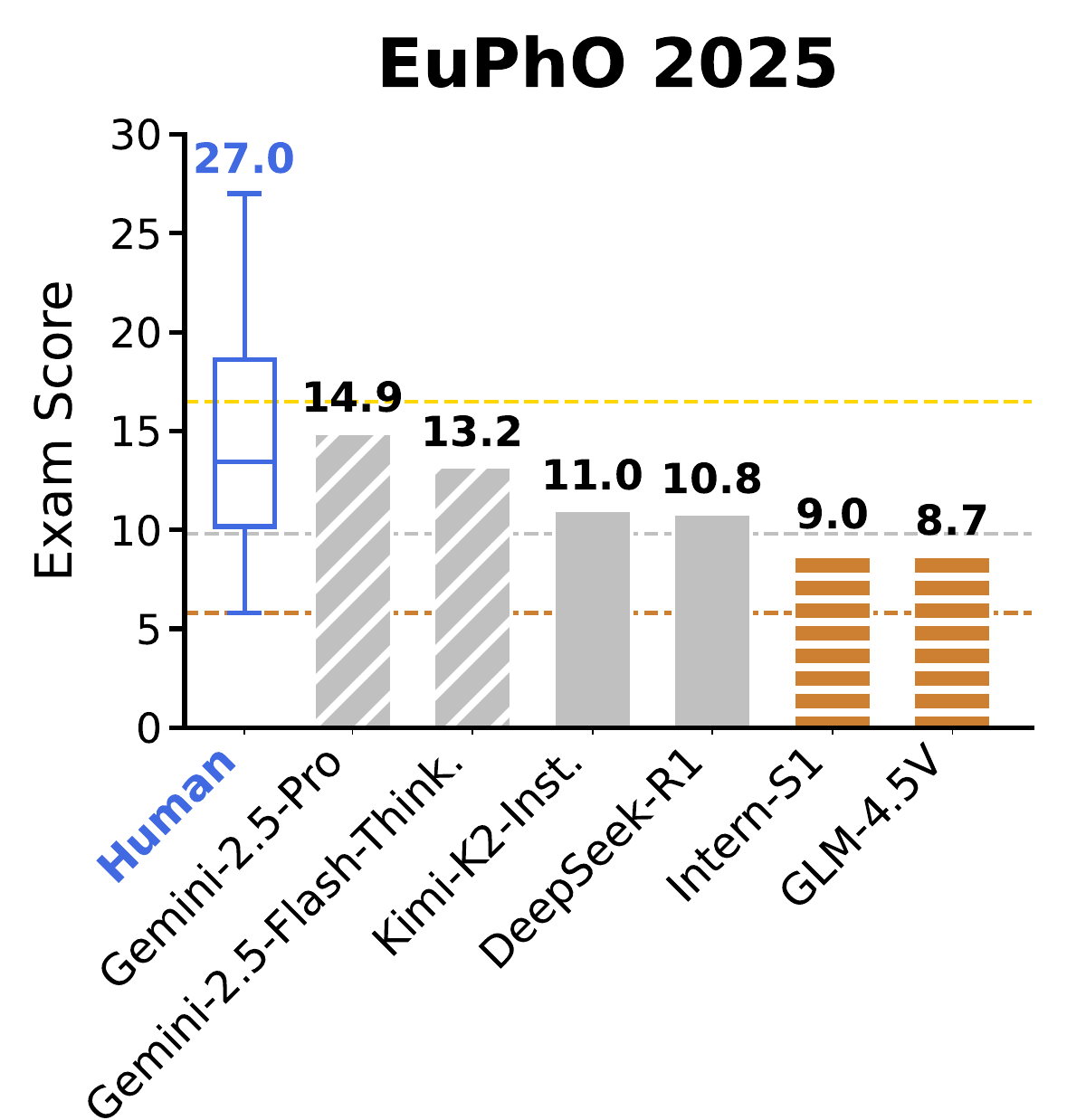}
    \includegraphics[width=0.328\linewidth]{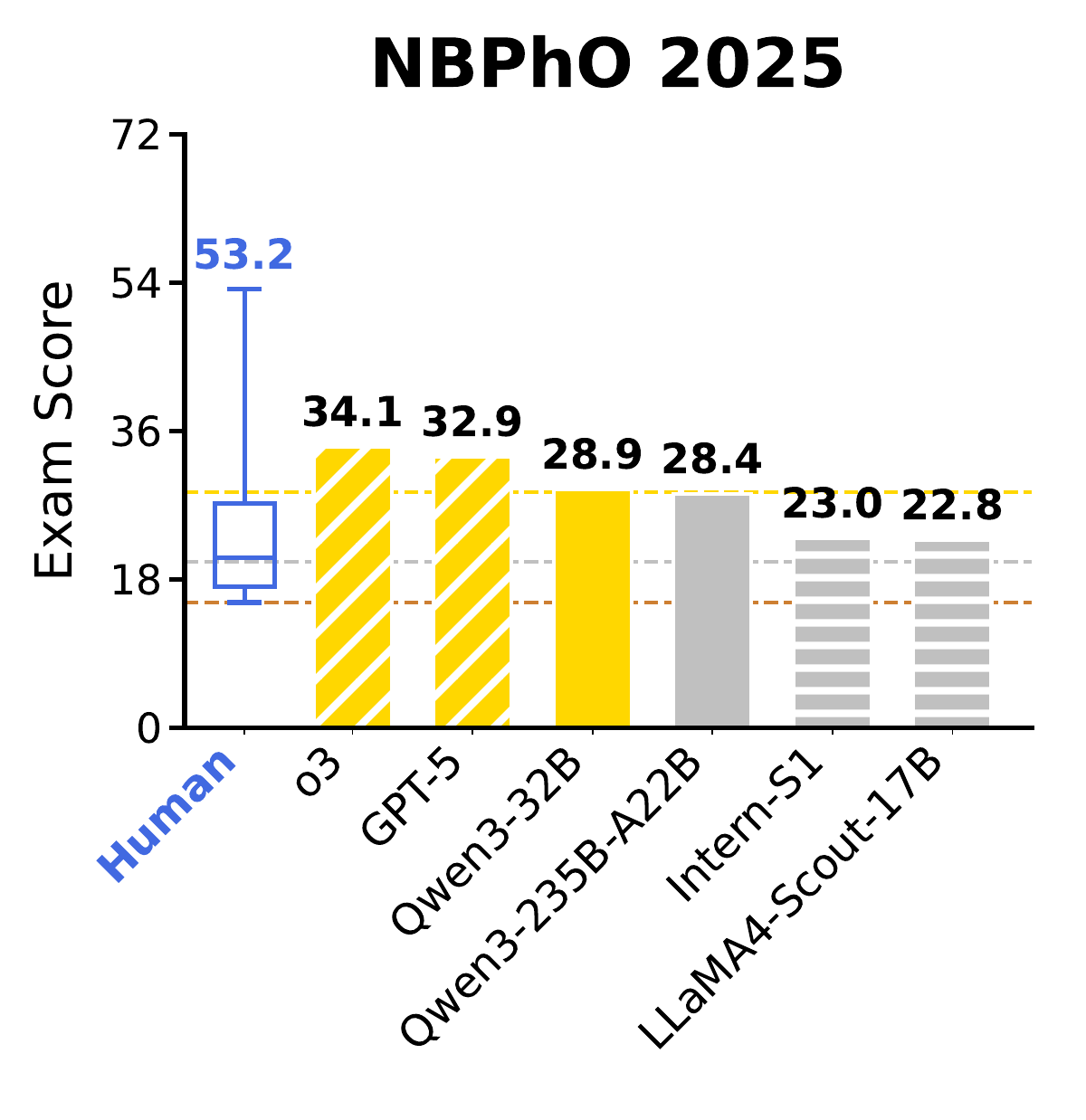}
    \includegraphics[width=0.328\linewidth]{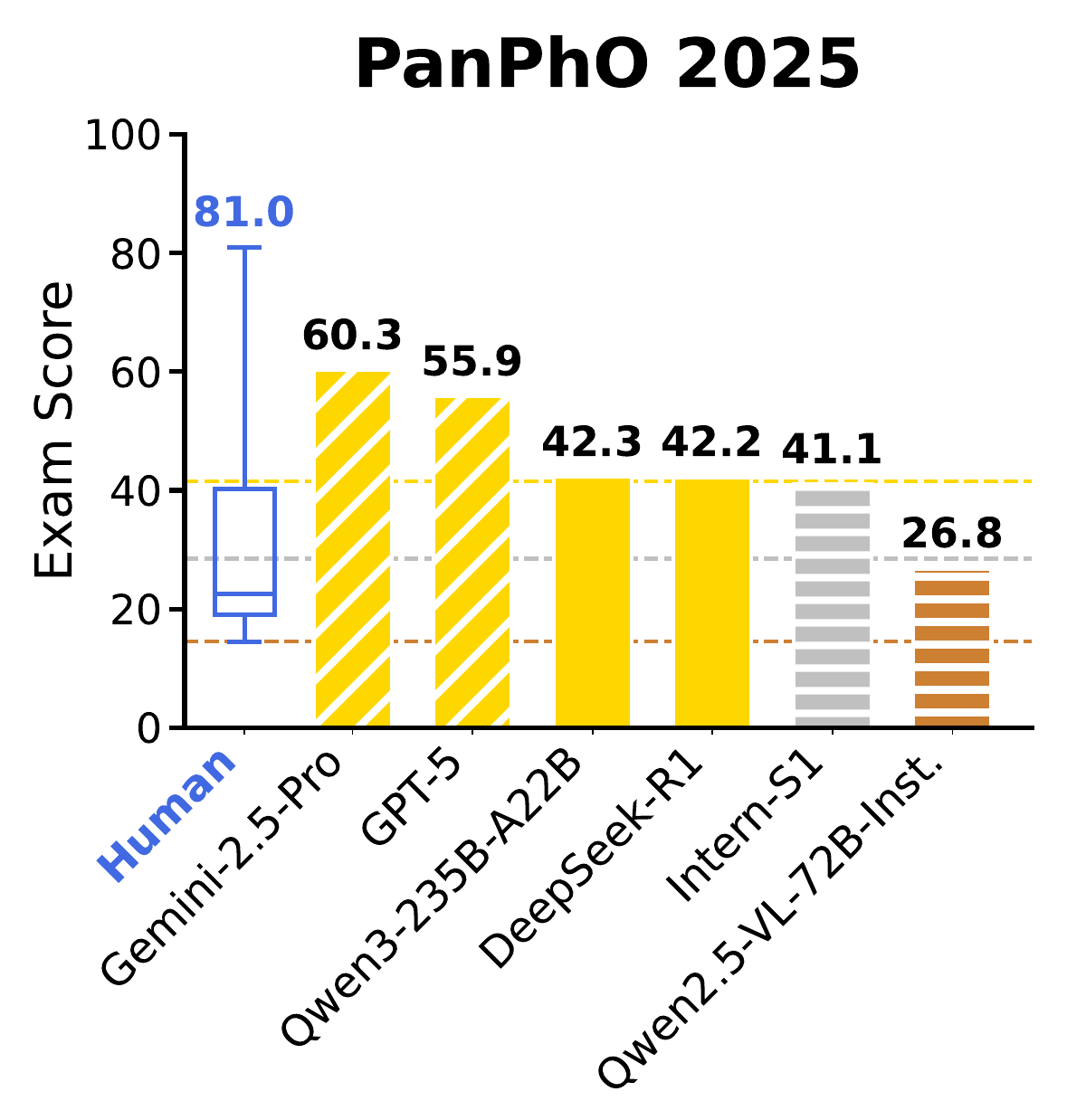}
    \includegraphics[width=0.328\linewidth]{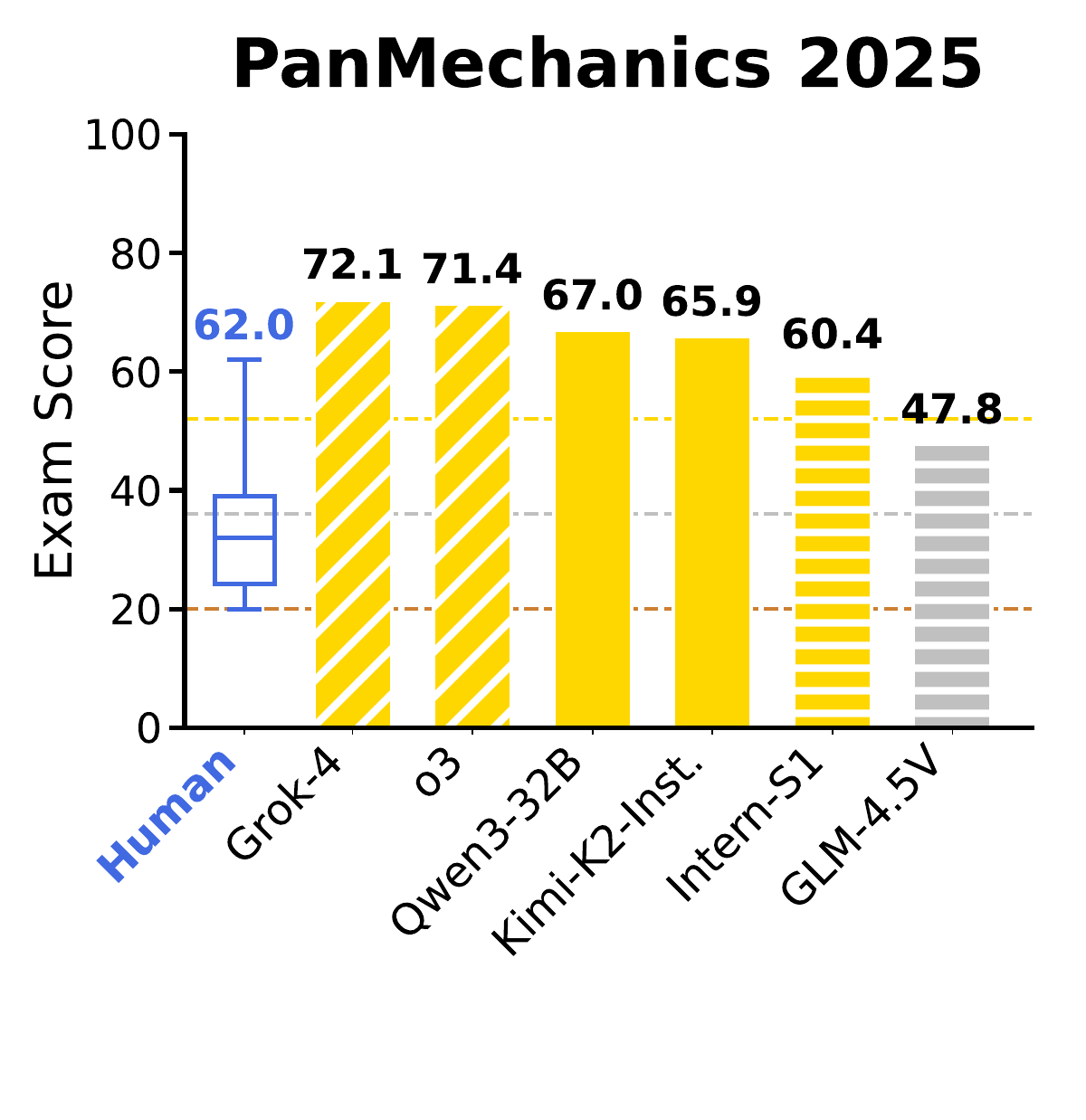}
    
    \caption{Exam scores of SOTA (M)LLMs and top-performing human medalists on the 2025 physics Olympiads. Each panel shows the score distribution of \textbf{human medalists} (boxplot), alongside bars for the top-2 models in closed-source MLLMs, open-source MLLMs, and open-source LLMs. Bars shaded \medalbox{Gold!50}{Gold}, \medalbox{Silver!70}{Silver} and \medalbox{Bronze!40}{Bronze} indicate scores above their respective medal thresholds. As shown, SOTA (M)LLMs often reach gold level but still fall short of the highest human scores in most exams.}
    \label{fig:score}
\end{figure}

\subsection{Can SOTA (M)LLMs Compete with Human Medalists in Physics Olympiads?}

\textbf{SOTA closed-source MLLMs remain unable to rival top-1 human contestants in most exams.}
As shown in Fig.~\ref{fig:score}, closed-source MLLMs frequently reached the gold threshold across Olympiads, yet consistently fell short of the very best human medalists in most exams. For example, Gemini-2.5-Pro achieved 22.7 points in IPhO-2025, the highest among all models, but 21 of the 37 human gold medalists still scored higher. In PanPhO-2025, its 60.3 points were far below the top human score of 81. These results show that while closed-source models can reliably ``reach the podium,'' they remain unable to rival the very top-performing students.

\textbf{SOTA open-source MLLMs usually reach bronze to silver levels, but still fall short of human gold medalists in most Olympiads.}  
As shown in Fig.~\ref{fig:score}, open-source models delivered respectable results in several contests, often scoring near the median of human medalists. For instance, in PanPhO-2025 Intern-S1 scored above 40 points, surpassing the median performance of human medalists, yet still below the gold threshold. By contrast, in more challenging Olympiads such as EuPhO-2025, which featured optics problems with variable figures, open-source MLLMs dropped back to the bronze level. These results illustrate both the steady progress of open-source models and the considerable gap that remains between them and the top-performing human contestants.

\textbf{Mechanics-focused Olympiads demonstrate the potential of open-source (M)LLMs to rival human gold medalists.}  
In PanMechanics-2025, which targets younger high-school students and places strong emphasis on classical mechanics, open-source (M)LLMs not only reached the gold threshold but in some cases even surpassed top human scores. For example, Kimi-K2-Instruct achieved 65.9 points compared to the top human score of 62.0. These results suggest that models are already approaching near-expert competence in mechanics, where reasoning is more deterministic and better aligned with training priors, while continuing to struggle in Olympiads that include optics problems, such as EuPhO-2025 and PanPhO-2024.


\subsection{Do visual inputs pose a real challenge for MLLMs’ physical reasoning?}

\begin{wrapfigure}{r}{0.45\textwidth}
\vspace{-4mm}
    \centering
    \includegraphics[width=0.45\textwidth]{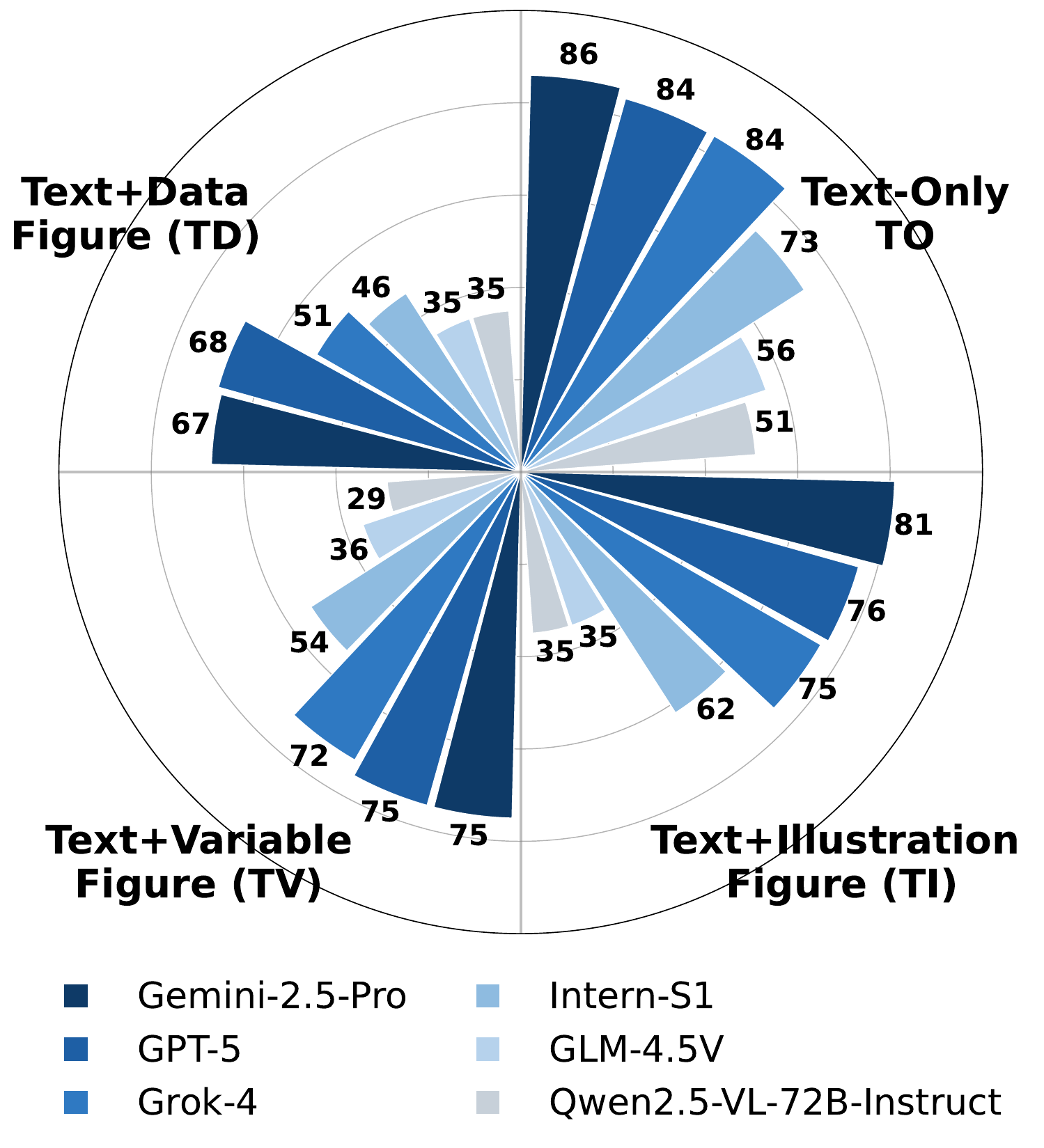}
    \vspace{-5mm}
    
    \caption{Comparison of mean normalized scores across four modality types.}
    \label{fig:modality_comparison}
    \vspace{-5mm}
\end{wrapfigure}

To investigate the impact of modality on performance, we categorize all problems into four types (Fig.~\ref{fig:problem_type}): Text-Only (TO), Text+Illustration Figure (TI), Text+Variable Figure (TV), and Text+Data Figure (TD). For each type, we report the \emph{Mean Normalized score} (MNS), defined as:
\begin{equation}
\label{eq:mean_normalized_score}
\small
\textbf{MNS($M$)} = \frac{1}{N_M} \sum_{Q \in M} \dfrac{\text{Exam Score}(Q)}{\text{Full Mark}(Q)} \times 100\%,
\end{equation}
where $M \in \{\text{TO}, \text{TI}, \text{TV}, \text{TD}\}$, $N_M$ is the number of questions in $M$, and $Q$ denotes a single question. 

As shown in Fig.~\ref{fig:modality_comparison}, diagram-based problems are consistently more difficult than text-only ones, leading to a sharp decline in mean normalized scores. For instance, the leading closed-source model, Gemini-2.5-Pro, the score reaches 86\% on TO but decreases progressively as visual complexity increases: 81\% on TI, 75\% on TV, and 67\% on TD. Grok-4 drops to 51\% on TD, highlighting persistent challenges in extracting numerical values from plots and interpreting functional plots such as peaks. Comparatively, open-source models fall further behind: the gap is visible on TO questions and widens significantly with visual input. On TV problems, GPT-5 achieves 75\%, while Qwen2.5-VL-72B-Instruct reaches only 29\%, showing its limitation in interpreting figures with complex variables.

Overall, these results highlight three major areas where the visual reasoning of MLLMs in physics problems can be further advanced: \textbf{(1)} interpreting illustration diagrams, \textbf{(2)} reasoning over variable-based graphs, and \textbf{(3)} accurately extracting quantitative information from data figures.


\subsection{Which Physics Fields Challenge MLLMs the Most?}

\begin{wraptable}{r}{0.45\textwidth}
\centering
\caption{Comparison of mean normalized scores (\%) across five major physics fields.}
\label{tab:results_physics_field}
\small
\resizebox{0.45\textwidth}{!}{%
\setlength{\tabcolsep}{1pt}
\begin{tabular}{lccccc}
\toprule
Physics Field & Mech. & Elec. & Ther. & Opt. & Mode. \\
\midrule
Gemini-2.5-Pro & 80 & 79 & 89 & 55 & 84 \\
GPT-5 & 80 & 72 & 82 & 52 & 79 \\
Grok-4 & 74 & 73 & 84 & 51 & 81 \\
\arrayrulecolor[gray]{0.7}\midrule\arrayrulecolor{black}
Intern-S1 & 63 & 67 & 64 & 41 & 64 \\
GLM-4.5V & 42 & 35 & 44 & 30 & 49 \\
Qwen2.5-VL-72B-Inst. & 37 & 32 & 48 & 23 & 50 \\
\bottomrule
\end{tabular}%
}
\end{wraptable}

Physics Olympiad problems span five fields: \textbf{Mechanics} (Mech.), \textbf{Electromagnetism} (Elec.), \textbf{Thermodynamics} (Ther.), \textbf{Optics} (Opt.), and \textbf{Modern Physics} (Mode.). We report mean normalized scores (Eq.~\ref{eq:mean_normalized_score}) for each field in Table~\ref{tab:results_physics_field}, leading to three key observations.
\textbf{(1) Optics is the most challenging field}, with all models scoring below 55\%. Its difficulty arises from two aspects: geometrical optics requires diagram interpretation, while wave optics demands precise symbolic derivations—both remain weak points for current MLLMs. 
\textbf{(2) Modern Physics tends to yield higher scores than Mechanics}, likely because the problems are less visually intensive and the concepts tested do not extend to university-level depth.
\textbf{(3) Mechanics and Electromagnetism show a clear source gap.} Closed-source models average above 70\%. Most open-source MLLMs remain below 45\%, with one exception: Intern-S1 narrows the gap, reaching 63–67\% in Mech./Elec.
Overall, optics is the most challenging, while closed-source models show steady gains.


\subsection{How Can (M)LLMs Achieve True Human-Level Physics Reasoning?}

Beyond theoretical performance, (M)LLMs face three structural limitations compared to human contestants. \textbf{(1) Multimodal ability}: LLMs lack the capacity to interpret diagrams, while MLLMs, though capable of reading figures, still struggle with variable-based and data-intensive figures. \textbf{(2) Generative ability}: physics Olympiads require not only interpreting diagrams but also sketching functional plots. Most (M)LLMs cannot generate such diagrams directly, and evaluating the quality of generated plots remains non-trivial. \textbf{(3) Embodied ability}: IPhO, APhO, and EuPhO include separate experimental exams, and NBPhO integrates theoretical and experimental problems within a single paper. Since most models lack embodiment and generative capabilities, all experimental and diagram-generation problems are excluded to ensure fair evaluation. As a result, the Full Mark (Model) in Table~\ref{tab:results} may be lower than the Full Mark (Human). A detailed overview of exam formats and scoring components is provided in Table~\ref{tab:exam_format} of Appendix~\ref{appendix:data_source}.

Reaching true human-level physics reasoning will thus require advances in three dimensions: \textbf{(1) multimodality}, for robust integration of textual and visual inputs; \textbf{(2) generation}, to produce diagrams and functional plots; and \textbf{(3) embodiment}, to support experimental reasoning and physical interaction. Without progress in these areas, even the most capable (M)LLMs will remain constrained relative to the holistic problem-solving abilities of human contestants.
\section{Conclusion and Future Work}

In this work, we presented \textsc{HiPhO}, the first benchmark designed to systematically evaluate (M)LLMs against top-performing students in high school physics Olympiads. Spanning 13 Olympiad exams from 2024–2025, \textsc{HiPhO} covers both international and regional contests. The problems include mixed modalities, spanning text-only to diagram-based, with fine-grained categorization. \textsc{HiPhO} is the first to combine answer-level and step-level grading based on official marking schemes, enabling rigorous, human-aligned evaluation. We compare model exam scores against human contestants and observe significant performance gaps. SOTA closed-source reasoning MLLMs achieve gold-level performance on most exams but still fall short of the very best human contestants. Most open-source MLLMs remain at the bronze level or below, while open-source LLMs obtain multiple golds. These findings highlight both the progress of (M)LLMs in physical reasoning and the substantial gap that remains before achieving human-level mastery.

Looking forward, \textsc{HiPhO} can be extended in two key directions. First, expanding coverage to include additional Olympiads such as the Chinese Physics Olympiad (CPhO) and the USA Physics Olympiad (USAPhO) would provide a broader and more representative benchmark for evaluating physics reasoning. Second, introducing a dynamic update mechanism that incorporates newly released Olympiad problems and human scores would ensure data freshness and benchmark integrity, while reducing risks of training contamination. By both broadening its scope and keeping the dataset continuously refreshed, future iterations of \textsc{HiPhO} can offer an even more comprehensive and reliable standard, further driving progress in multimodal physical reasoning.

\subsubsection*{Acknowledgments}

This work was supported by a locally commissioned task from the Shanghai Municipal Government. 
We also gratefully acknowledge the support of the APhO Committee.

\bibliography{ref}
\bibliographystyle{iclr2025_conference}

\newpage
\appendix

\begin{center}
    \textbf{\Large Supplemental Material of \textsc{HiPhO} Benchmark}
\end{center}

\section{Details of \textsc{HiPhO} Benchmark}
\label{appendix:benchmark}

\subsection{Data Source and Exam Format}
\label{appendix:data_source}

\textsc{HiPhO} covers seven types of high school physics Olympiads, from international to regional exams.

\begin{itemize}
    \item \textbf{IPhO (International Physics Olympiad):}  
    The most prestigious global physics Olympiad for high school students since 1967, featuring challenging theoretical and experimental exams.

    \item \textbf{APhO (Asian Physics Olympiad):} 
    A regional contest launched in 2000 for Asian and Oceanian students, structured similarly to IPhO with both theory and laboratory components.

    \item \textbf{EuPhO (European Physics Olympiad):}  
    Established in 2017, a multi-day competition for European students that emphasizes creative problem solving in both theory and experiment.

    \item \textbf{NBPhO (Nordic-Baltic Physics Olympiad):}  
    A regional contest among Nordic and Baltic countries, primarily theoretical but also including experimental problems.

    \item \textbf{PanPhO (Pan Pearl River Delta Physics Olympiad):}  
    An invitational competition for top schools from the Pearl River Delta and neighboring regions in China.

    \item \textbf{PanMechanics (Pan Pearl River Delta Mechanics Test):}  
    A specialized subset of PanPhO focusing solely on mechanics, typically offered as a shorter, single-field exam.

    \item \textbf{F=MA:}  
    A U.S. mechanics contest organized by the American Association of Physics Teachers, serving as the entry test for the U.S. Physics Olympiad (USAPhO).
\end{itemize}
\vspace{-3mm}

\begin{table}[H]
\centering
\caption{Source links for the 13 collected physics Olympiad exams in the \textsc{HiPhO} benchmark.}
\label{tab:data_source}
\small
\setlength{\tabcolsep}{8.5pt}
\begin{tabular}{l|lll|lll}
\toprule
Year & \multicolumn{3}{c|}{2025} & \multicolumn{3}{c}{2024} \\ 
Physics Olympiad & Problem & Solution & Results & Problem & Solution & Results \\  \midrule
IPhO & \href{https://ipho.olimpicos.net/pdf/IPhO_2025_Q1.pdf}{Q1}, \href{https://ipho.olimpicos.net/pdf/IPhO_2025_Q2.pdf}{Q2}, \href{https://ipho.olimpicos.net/pdf/IPhO_2025_Q3.pdf}{Q3} & \href{https://ipho.olimpicos.net/pdf/IPhO_2025_S1.pdf}{S1}, \href{https://ipho.olimpicos.net/pdf/IPhO_2025_S2.pdf}{S2}, \href{https://ipho.olimpicos.net/pdf/IPhO_2025_S3.pdf}{S3} & \href{https://cdn.prod.website-files.com/664df830da8ff5d22656764b/68834cd4e6e698960f2c6e1a_final_ranks_ipho2025.pdf}{results} & \href{https://ipho.olimpicos.net/pdf/IPhO_2024_Q1.pdf}{Q1}, \href{https://ipho.olimpicos.net/pdf/IPhO_2024_Q2.pdf}{Q2}, \href{https://ipho.olimpicos.net/pdf/IPhO_2024_Q3.pdf}{Q3} & \href{https://ipho.olimpicos.net/pdf/IPhO_2024_S1.pdf}{S1}, \href{https://ipho.olimpicos.net/pdf/IPhO_2024_S2.pdf}{S2}, \href{https://ipho.olimpicos.net/pdf/IPhO_2024_S3.pdf}{S3} & \href{https://www.ipho2024.ir/page/10029}{results} \\
APhO & \href{https://www.apho2025.sa/media/h5aipfl1/theory_questions_apho_2025.pdf}{problem} & \href{https://www.apho2025.sa/media/zlzlx5zy/theory_solution_apho_2025.pdf}{solution} & \href{https://www.apho2025.sa/media/4mkc32ln/apho_2025_\%D9\%86\%D8\%B4\%D8\%B1-\%D8\%A7\%D9\%84\%D9\%86\%D8\%AA\%D8\%A7\%D8\%A6\%D8\%AC-1.pdf}{results} & \multicolumn{3}{c}{Human results are unavailable} \\
EuPhO & \href{https://eupho.ee/wp-content/uploads/2025/06/EuPhO_2025_Theory_ENG.pdf}{problem} & \href{https://eupho.ee/wp-content/uploads/2025/06/EuPhO25_Theory_Solutions_20250616_1120.pdf}{solution} & \href{https://eupho.ee/wp-content/uploads/2025/08/EuPhO-2025-results-070825.pdf}{results} & \href{https://eupho.ee/wp-content/uploads/2024/07/EuPhO_2024_theory_final.pdf}{problem} & \href{https://eupho.ee/wp-content/uploads/2024/07/EuPhO_2024_theory_sol.pdf}{solution} & \href{https://eupho.ee/wp-content/uploads/2024/07/EuPhO-2024-results-resultsforweb.pdf}{results} \\
NBPhO & \href{https://nbpho.ee/wp-content/uploads/2025/04/NBPhO_2025_problems_V1.0.pdf}{problem} & \href{https://nbpho.ee/wp-content/uploads/2025/04/00_NBPhO_2025_solutions_v1.0.pdf}{solution} & \href{https://nbpho.ee/wp-content/uploads/2025/04/NBPhO-2025-grading-Final-for-web.pdf}{results} & \href{https://nbpho.ee/wp-content/uploads/2024/04/NBPhO_2024_problems_v1.0.pdf}{problem} & \href{https://nbpho.ee/wp-content/uploads/2024/04/NBPhO_2024_solutions_v0.98.pdf}{solution} & \href{https://nbpho.ee/wp-content/uploads/2024/04/NBPhO-2024-final-results.pdf}{results} \\
PanPhO & \href{https://panpearl-phys.hkust.edu.hk/files/result_2025/PanPearl2025-Paper\%201.pdf}{Q1},  \href{https://panpearl-phys.hkust.edu.hk/files/result_2025/PanPearl2025-paper%202.pdf}{Q2} & \href{https://panpearl-phys.hkust.edu.hk/files/result_2025/PanPearl2025-Paper\%201-sol.pdf}{S1}, \href{https://panpearl-phys.hkust.edu.hk/files/result_2025/PanPearl2025-Paper%202-sol.pdf}{S2} & \href{https://panpearl-phys.hkust.edu.hk/files/result_2025/HK\%20result.pdf}{results} & \href{https://panpearl-phys.hkust.edu.hk/files/pastpaper/Pan\%20Pearl\%202024\%20paper\%201\%20-\%20question.pdf}{Q1}, \href{https://panpearl-phys.hkust.edu.hk/files/pastpaper/PanPearl2024%20paper%202_question.pdf}{Q2} & \href{https://panpearl-phys.hkust.edu.hk/files/pastpaper_sol/Pan\%20Pearl\%202024\%20paper\%201\%20-\%20sol.pdf}{S1}, \href{https://panpearl-phys.hkust.edu.hk/files/pastpaper_sol/PanPearl2024%20paper%202_solution.pdf}{S2} & \href{https://panpearl-phys.hkust.edu.hk/files/result_2024/HK\%20result.pdf}{results} \\
PanMechanics & \href{https://panpearl-phys.hkust.edu.hk/files/result_2025/Pan_PhO_paper_2025_20250115.pdf}{problem} & \href{https://panpearl-phys.hkust.edu.hk/files/result_2025/PanPearl2025-mech-sol.pdf}{solution} & \href{https://panpearl-phys.hkust.edu.hk/files/result_2025/2025\%20HK\%20mechanics\%20result.pdf}{results} & \href{https://panpearl-phys.hkust.edu.hk/files/mechanics/Pan_PhO_paper_2024_sc.pdf}{problem} & \href{https://panpearl-phys.hkust.edu.hk/files/mechanics/Pan_PhO_paper_2024_sc_solution.pdf}{solution} & \href{https://panpearl-phys.hkust.edu.hk/files/result_2024/HK_mechanics_result_2024.pdf}{results} \\
F=MA & \href{https://www.aapt.org/physicsteam/upload/FMA-exam.pdf}{problem} & \href{https://www.aapt.org/physicsteam/upload/FMA_solutions.pdf}{solution} & \href{https://www.aapt.org/physicsteam/2025/upload/2025_F_ma_results.pdf}{results} & \href{https://www.aapt.org/physicsteam/upload/2024_F-ma_Exam.pdf}{problem} & \href{https://www.aapt.org/physicsteam/upload/F-ma-2024-Solutions_v2.pdf}{solution} & \href{https://www.aapt.org/physicsteam/2024/upload/2024_F_ma_results.pdf}{results} \\
\bottomrule
\end{tabular}
\end{table}


Physics Olympiads adopt diverse exam formats. As shown in Table~\ref{tab:exam_format}, IPhO, APhO, and EuPhO include a separate 20-point experimental exam, while NBPhO combines theoretical and experimental components into a single paper (72-point total). PanPhO, PanMechanics, and F=MA include only theoretical problems.
Since most (M)LLMs lack embodied experimental and diagram-drawing capabilities, we exclude all experimental and diagram-generation questions for fair and consistent evaluation. The \textbf{Full Mark (Model)} in Table~\ref{tab:results} therefore refers only to the \textbf{theoretical score} of the theoretical exam.

Importantly, we introduce for the first time a step-level score based on official marking schemes. Among the 13 exams, 7 provide \textbf{official marking schemes}, and 4 support \textbf{multiple solutions}. This alignment improves fairness and ensures consistency with human grading standards.
\vspace{-1mm}

\begin{table}[H]
\centering
\caption{Exam formats and scoring components of physics Olympiads in the \textsc{HiPhO} benchmark.}
\label{tab:exam_format}
\small
\resizebox{\textwidth}{!}{%
\setlength{\tabcolsep}{0.1pt}
\begin{tabular}{
    l 
    *{13}{>{\centering\arraybackslash}p{0.065\textwidth}} 
}
\toprule
\textbf{Physics Olympiad} & \multicolumn{2}{c}{\textbf{IPhO}} & \textbf{APhO} & \multicolumn{2}{c}{\textbf{EuPhO}} & \multicolumn{2}{c}{\textbf{NBPhO}} & \multicolumn{2}{c}{\textbf{PanPhO}} & \multicolumn{2}{c}{\textbf{PanMechanics}} & \multicolumn{2}{c}{\textbf{F=MA}}  \\
\textbf{Year} & 2025 & 2024 & 2025 & 2025 & 2024 & 2025 & 2024 & 2025 & 2024 & ~~2025 & 2024 & 2025 & 2024  \\ \midrule
Theoretical Exam Score & 30 & 30 & 30 & 30 & 30 & 72 & 72 & 100 & 100 & 100 & 100 & 25 & 25 \\
\quad - \textbf{Theoretical Score} & 29.4 & 29.3 & 30 & 29 & 28 & 43.5 & 50 & 100 & 98 & 100 & 100 & 25 & 25 \\
\quad - Diagram Score & 0.6 & 0.7 & -- & 1 & 2 & 4.5 & -- & -- & 2 & -- & -- & -- & -- \\
\quad - Experimental Score & -- & -- & -- & -- & -- & 24 & 22 & -- & -- & -- & -- & -- & -- \\
Experimental Exam Score & 20 & 20 & 20 & 20 & 20 & -- & -- & -- & -- & -- & -- & -- & -- \\
\arrayrulecolor[gray]{0.7}\midrule\arrayrulecolor{black}
Official Marking Scheme & \ding{51} & \ding{51} & \ding{51} & \ding{51} & \ding{51} & \ding{51} & \ding{51} & \ding{55} & \ding{55} & \ding{55} & \ding{55} & \ding{55} & \ding{55}  \\
Multiple Solutions & \ding{55} & \ding{55} & \ding{51} & \ding{55} & \ding{51} & \ding{51} & \ding{51} & \ding{55} & \ding{55} & \ding{55} & \ding{55} & \ding{55} & \ding{55} \\
\bottomrule
\end{tabular}%
}
\end{table}


\subsection{Data Processing Pipeline}
\label{appendix:data_processing}

\begin{figure}[t]
    \centering
    \includegraphics[width=1\linewidth]{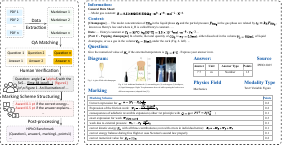}
    \vspace{-4mm}
    
    \caption{High-quality data processing pipeline (left) and an example of \textsc{HiPhO} data (right).}
    \label{fig:data_processing_pipeline}
\end{figure}

We implemented a structured, multi-stage data processing pipeline, as illustrated in Fig.~\ref{fig:data_processing_pipeline} (left).

\begin{itemize}
    \item \textbf{Data Extraction.} Official Olympiad exam papers in PDF format were processed with OCR-based tools. Text blocks were reflowed into markdown for easier parsing and editing.
    
    \item \textbf{QA Matching.} Questions were aligned with their official answers by matching indices and numbering patterns, ensuring that every problem statement had a corresponding answer entry.
    
    \item \textbf{Human Verification.} Human experts carefully reviewed all extracted content. This included correcting OCR misrecognitions and ensuring textual consistency with the original exam.  
    
    \item \textbf{Marking Scheme Structuring.} For exams with marking schemes, step-level scoring criteria were systematically extracted, and then reformatted into standardized, model-readable rules.
    
    \item \textbf{Post-Processing.} Verified QAs underwent additional refinement:  
    \begin{itemize}
        \item \textit{Context Completion:} Long problem stems and necessary background information were consolidated to provide self-contained contexts.  
        \item \textit{Subquestion Structuring:} Interdependent subparts were explicitly separated and re-labeled to reflect the intended order of solution.  
        \item \textit{Unit Specification:} Required physical units were clarified in the problem statement to minimize ambiguity and avoid unfair penalization.  
    \end{itemize}
\end{itemize}

As shown in Fig.~\ref{fig:data_processing_pipeline} (right), the finalized dataset is organized into a unified \texttt{json} format:

\begin{itemize}
    \item \textbf{Information}: Physics constants table on the exam cover page (if available), e.g., $g=9.8~m/s^2$.  
    \item \textbf{Context}: Problem stem content, including introductory descriptions or preceding subparts.  
    \item \textbf{Question}: Specific question text, with multiple subquestions explicitly split into (1), (2), etc.  
    \item \textbf{Diagram}: Path to required figures or diagrams associated with the question.  
    \item \textbf{Marking}: Step-level scoring points in the unified form “Award xx pt if the answer …”.  
    \item \textbf{Answer}: Ground-truth solution including value, units, answer type, and allocated points.  
    \item \textbf{Physics Field}: The physics domain of the problem (Mechanics, Electromagnetism, Thermodynamics, Optics, Modern Physics).  
    \item \textbf{Modality Type}: The modality type of the problem (Text-Only, Text+Illustration Figure, Text+Variable Figure, Text+Data Figure).  
    \item \textbf{Source}: The corresponding Olympiad exam, e.g., IPhO 2025.  
\end{itemize}


\section{Details of Evaluation Framework}
\label{appendix:evaluation}

\subsection{Inference Prompt}
\label{appendix:inference_prompt}

To ensure consistent and reproducible evaluation across models, we adopt a compact instruction template that: (i) enforces LaTeX-formatted math, (ii) separates full reasoning from the final answer using \texttt{\textless think\textgreater} and \texttt{\textless answer\textgreater} tags, and (iii) standardizes multi-part and multiple-choice outputs with boxed answers.

To align with the original language of each Olympiad exam, we use English prompts for English-language exams and Chinese prompts for Chinese-language exams. Specifically, the language settings are as follows:
\begin{itemize}
\item \textbf{English Exams:} IPhO, APhO, EuPhO, NBPhO, PanPhO, F=MA
\item \textbf{Chinese Exam:} PanMechanics
\end{itemize}

For completeness, we provide both the English and Chinese prompt templates below.
\vspace{10mm}

\begin{promptbox}[Inference Prompt for English Exams]
You are participating in a high school physics Olympiad exam.\\

Please read the following question carefully and provide a clear, step-by-step solution with full reasoning.\\

Instructions:\\
1. Use LaTeX to format all variables, equations, and calculations.\\
2. Enclose your full reasoning process within \textless think\textgreater  \textless /think\textgreater ~ tags.\\
3. Provide the final answer within \textless answer\textgreater  \textless /answer\textgreater ~ tags, using the format of [\textbackslash boxed\{answer\}]. Do not include units inside the box.\\
4. For multiple sub-questions, list the answers in order using the format: [\textbackslash boxed\{answer1\}, \textbackslash boxed\{answer2\}, ...].\\
5. For multiple-choice questions, provide the final selected option(s) in the boxed answer instead of the calculation result (e.g., [\textbackslash boxed\{A\}]).\\

Example of Output:\\
\textless think\textgreater\\
Step 1: Analyze the problem... Step 2: Apply the relevant equations...\\
\textless /think\textgreater \\
\textless answer\textgreater

[\textbackslash boxed\{A\}, \textbackslash boxed\{3.2\}]

\textless /answer\textgreater\\

Useful information (formulas, constants, units, if applicable):\\
\{information\}\\

Context (if applicable):\\
\{context\}

Question (Answer only the question stated below):\\
\{question\}
\end{promptbox}

\newpage
\begin{CJK*}{UTF8}{gbsn}  
\begin{promptbox}[Inference Prompt for Chinese Exams]
你正在参加高中物理竞赛。\\

请仔细阅读下列题目，结合上下文信息，详细推导并给出清晰、有条理的解题步骤与完整的逻辑推理过程。\\

作答要求：\\
1. 所有物理量、公式和计算过程须使用 LaTeX 格式书写。\\ 
2. 将完整的推理过程用 \textless think\textgreater 和 \textless /think\textgreater 标签括起来。\\
3. 将最终答案置于 \textless answer\textgreater  和 \textless /answer\textgreater  标签中，答案格式为 [\textbackslash boxed\{答案\}]，方框内不包含单位。\\
4. 对于包含多个小问的题目，按顺序列出所有答案，格式为：[\textbackslash boxed\{答案1\}, \textbackslash boxed\{答案2\}, ...]。\\
5. 对于选择题，请在答案的方框中给出最终选择的选项，而不是计算结果（例如：[\textbackslash boxed\{A\}]）。\\

输出示例：\\
\textless think\textgreater \\
第一步：分析问题... 第二步：运用相关公式...\\
\textless /think\textgreater \\
\textless answer\textgreater 

[\textbackslash boxed\{A\}, \textbackslash boxed\{3.2\}]

\textless /answer\textgreater \\

可用信息（如物理公式、常数、单位等）：\\
\{information\} \\

背景信息（如有）：\\
\{context\} \\

题目内容（仅回答以下问题）：\\
\{question\}
\end{promptbox}
\end{CJK*}


\subsection{Answer-Level Coarse-Grained Score}
\label{appendix:answer_level}

The answer-level coarse-grained evaluation pipeline extends the PHYSICS framework \citep{2025physics2} and follows a structured sequence to assess whether the model's final answers are correct. The process comprises the following key steps:

\begin{itemize}
    \item \textbf{Answer Extraction.} Final answers are extracted from the model's solution using automatic parsing of boxed expressions (e.g., \verb|\boxed{...}|). This ensures that evaluation targets the model's intended outputs, while ignoring intermediate reasoning steps.

    \item \textbf{Rule-Based Matching.} A rule-based math-verifier \citep{math_verify} compares the extracted answers with the ground-truth answers. It performs correct matching on either numeric or symbolic expressions, while also accounting for units and answer types.

    \item \textbf{Model-Based Verification.} If no exact match is found, a second-pass verification is performed using a powerful judge model (Gemini-2.5-Flash). This model compares the model's answer and the ground-truth answer to determine physical or mathematical equivalence, thereby reducing false negatives caused by alternative expressions, equivalent transformations, or formatting inconsistencies.

    \item \textbf{Multi-Part Matching Logic.} For problems consisting of multiple sub-questions (e.g., labeled as (1), (2), etc.), the intended answer order is explicitly defined by the problem statement. To reflect this, we extract boxed answers from the model's response in reverse order and compare them sequentially with the ground-truth, following the original question structure.
\end{itemize}

Unlike the original PHYSICS's evaluation framework, which relied on a fine-tuned lightweight 8B verifier, our pipeline leverages Gemini-2.5-Flash for model-based verification. This substitution substantially improves the reliability of equivalence judgments. The corresponding evaluation prompt is adapted from the PHYSICS design \citep{2025physics2}, as shown below.

\vspace{6mm}

\begin{promptbox}[Prompt for Answer-level Model-based Evaluation]
You are a diligent and precise assistant tasked with evaluating the correctness of responses. You will receive a question, an output sentence, and the correct answer. Your task is to determine if the output sentence accurately answers the question based on the provided correct answer. Respond with either [Correct] or [Incorrect].\\

Special considerations:\\
1. \textbf{Multiple Answers:} If the output contains multiple answers, evaluate whether later answers modify or correct earlier ones. In such cases, compare the final answer with the correct answer. If the final answer is unclear or incorrect, respond with [Incorrect].\\
2. \textbf{Mathematical Problems:} If the formats differ but the answers are mathematically equivalent such as 256/55=4.65, respond with [Correct].\\
3. \textbf{Physics Problems:} If the values match such as 3=3 GHz, respond with [Correct].\\
4. \textbf{Explicit Options:} If the question provides explicit candidate answers, the output will be considered correct if it clearly indicates the correct option's code or the correct option's content.\\
5. \textbf{No Explicit Options:} If the question does not provide explicit options, the output must align
with the correct answer in content and meaning to be considered [Correct].\\

Question: \{problem\} \\
Output sentence: \{given\_answer\} \\
Correct answer: \{ground\_truth\} \\

Final Instruction:\\
You must respond with exactly one of the following: [Correct] or [Incorrect]. \\
Do NOT include any explanation, reasoning, or additional text. \\
Any deviation from this format (even a single word) will be considered INVALID.\\

Judgement:
\end{promptbox}


\subsection{Step-Level Fine-Grained Score}
\label{appendix:step_level}

The step-level fine-grained evaluation measures the model's reasoning quality by comparing its solution steps against detailed criteria from official marking schemes, with each criterion representing a specific conceptual, physical, or mathematical step. The evaluation follows the steps below:

\begin{itemize}
    \item \textbf{Parse Marking Criteria.} Each problem is accompanied by a list of step-level marking points, such as “Applies energy conservation correctly” or “Derives the correct force expression.” These marking points are parsed from the dataset and include descriptions and assigned partial scores.

    \item \textbf{Model-Based Step Scoring.} For each marking point, the model's solution is independently assessed using Gemini-2.5-Flash, a strong judge model. A dedicated prompt is constructed to instruct the model to evaluate whether the student's solution satisfies the given criterion and to return a numerical score (e.g., 1.0, 0.5, or 0.0) accordingly. This approach enables semantic and physical equivalence checking beyond surface-level matching.

    \item \textbf{Score Aggregation.} The scores assigned for each marking point are aggregated to compute the final fine-grained score for the question. These scores are stored alongside their corresponding criteria for detailed feedback and analysis. For EuPhO and NBPhO problems with multiple official marking schemes, we take the highest score across schemes, reflecting the fact that solutions may follow different valid approaches.
\end{itemize}

Compared to commonly used answer-level evaluation, this fully model-based step-level evaluation provides a more accurate and context-aware assessment of the model's reasoning. The fine-grained scoring closely aligns with how human graders assign partial credit in real Olympiad exams, making it one of the core innovations of our benchmark and a more faithful reflection of true exam performance. The prompt used for step-level model-based evaluation is shown below.
\vspace{6mm}

\begin{promptbox}[Prompt for Step-level Model-based Evaluation]
You are an expert physics competition grader. Evaluate the student's solution against the specific grading criterion.\\

Physics Problem:\\
\{question\}\\

Student's Solution:\\
\{prediction\}\\

Grading Criterion:\\
\{criterion['description']\}\\

Instructions:\\
1. Analyze the student's solution for physics concepts, mathematical derivations, and calculations.\\
2. Award points strictly according to the criterion.\\
3. Consider both conceptual understanding and technical accuracy.\\

Critical: \\
1. You MUST respond with ONLY a single number (e.g., 1.0, 0.5, 0.0).\\ 
2. NO explanations, NO text, NO reasoning - JUST THE NUMBER.\\
3. If you provide any text other than the number, your response will be invalid.\\

Score:
\end{promptbox}


\subsection{Answer-Level vs. Step-Level vs. Human Expert Scores}
\label{appendix:model_score_human}

As shown in Table~\ref{tab:evaluation_score}, we compared grading outcomes across different graders. Step-level scores are typically higher than answer-level scores in the final exam results, reflecting the partial credit awarded for intermediate steps. Moreover, when used as a grader, the more powerful model Gemini-2.5-Flash produces results much closer to those of human experts, demonstrating its greater accuracy in evaluating exam performance, particularly at the step level.

\begin{table}[H]
\centering
\caption{Comparison of human expert grading with answer- and step-level scores, using Gemini-2.5-Flash or GPT-4o as graders, on the IPhO 2024 exam (single inference run example).}
\label{tab:evaluation_score}
\small
\setlength{\tabcolsep}{4.3pt}
\begin{tabular}{l|c|cc|cc|c|cc|cc}
\toprule
Contestant & \multicolumn{5}{c|}{\textbf{Gemini-2.5-Pro}} & \multicolumn{5}{c}{\textbf{Claude-4-Sonnet}} \\ \midrule
\multirow{2}{*}{Grader} & Human & \multicolumn{2}{c|}{Gemini-2.5-Flash} & \multicolumn{2}{c|}{GPT-4o} 
 & Human & \multicolumn{2}{c|}{Gemini-2.5-Flash} & \multicolumn{2}{c}{GPT-4o} \\
 & Expert & answer & step & answer & step
 & Expert & answer & step & answer & step \\
\midrule
Q1-A & 3.0 & 1.3 & 3.0 & 1.3 & 2.6 & 2.7 & 2.2 & 2.8 & 2.2 & 2.4 \\
Q1-B & 4.4 & 4.4 & 6.0 & 4.4 & 4.4 & 3.7 & 1.6 & 3.8 & 1.0 & 1.8 \\
Q2-A & 5.2 & 5.2 & 4.7 & 5.6 & 3.4 & 4.5 & 3.7 & 3.7 & 3.7 & 3.6 \\
Q2-B & 4.3 & 1.9 & 3.8 & 1.9 & 2.2 & 3.4 & 1.5 & 2.8 & 1.5 & 2.1 \\
Q3-A & 4.1 & 3.7 & 4.3 & 3.7 & 3.6 & 2.2 & 1.4 & 2.2 & 1.4 & 2.3 \\
Q3-B & 3.8 & 2.3 & 2.8 & 2.3 & 2.7 & 2.2 & 1.9 & 3.0 & 1.3 & 2.1 \\
\midrule
Exam Score & 24.8 & 18.8 & 24.6 & 19.2 & 18.9 & 18.7 & 12.3 & 18.3 & 11.1 & 14.3 \\
Diff. vs Human & -- & {\color{deepblue}-6.0} & {\color{deepblue}-0.2} & {\color{deepblue}-5.6} & {\color{deepblue}-5.9} & -- & {\color{deepblue}-6.4} & {\color{deepblue}-0.4} & {\color{deepblue}-7.6} & {\color{deepblue}-4.4} \\
\bottomrule
\end{tabular}
\end{table}


\newpage
\section{Details of Experiments and Results}
\label{appendix:experiment}

\subsection{Experimental Setup}
\label{appendix:experimental_setup}

\textbf{Hyperparameter Settings.} We adopt VLMEvalKit \citep{2024vlmevalkit} as the evaluation framework for benchmarking MLLMs. To reduce randomness and improve the reliability of evaluation, each problem is tested using \textbf{eight independent inference runs} at a \textbf{temperature of 0.6}; its score is averaged across runs, and the exam score is the sum of these averages. This setup helps reduce score fluctuations caused by randomness in the model's responses. In addition, the \textbf{maximum token limit} is set in reference to the largest value permitted by each model, helping prevent output truncation and ensuring that responses are complete and valid.

\textbf{Evaluated Models.} To assess the physics reasoning capabilities of state-of-the-art (M)LLMs relative to top-performing human contestants, we evaluate 30 representative models, including 11 closed-source MLLMs, 11 open-source MLLMs, and 8 open-source LLMs, based on the following criteria:
\textbf{(1) Recency:} Most models were released after April 2025, with the newest launched in August 2025 (e.g., GPT-5).
\textbf{(2) Diversity:} The selection includes both closed- and open-source models, covering reasoning-specialized and general-purpose architectures from a wide range of developers.
\textbf{(3) Model Scale:} A range of model sizes—small, medium, and large—is included to enable performance comparisons by scale.
The full list of evaluated models is provided in Table~\ref{tab:model}.

\begin{table}[H]
\vspace{-1mm}
\centering
\caption{Evaluated (M)LLMs grouped by source type, developer, and capability.}
\label{tab:model}
\small
\setlength{\tabcolsep}{4.7pt}
\begin{tabular}{llll}
\toprule
\textbf{Source} & \textbf{Developer} & \textbf{Reasoning Models} & \textbf{Chat Models} \\
\midrule
\multirow{5}{*}{\makecell[l]{\textbf{Closed-source} \\ \textbf{MLLMs}}}
& OpenAI & GPT-5, o3, o4-mini (high), o4-mini & GPT-4o \\
& Google & Gemini-2.5-Pro, Gemini-2.5-Flash-Thinking & -- \\
& Anthropic & Claude-4-Sonnet-Thinking & Claude-4-Sonnet \\
& xAI & Grok-4 & -- \\
& Mistral & -- & Mistral-Medium-3 \\
\arrayrulecolor[gray]{0.7}\midrule\arrayrulecolor{black}
\multirow{6}{*}{\makecell[l]{\textbf{Open-source} \\ \textbf{MLLMs}}}
& Zhipu AI & GLM-4.5V & -- \\
& Shanghai AI Lab & Intern-S1 & InternVL3 Series \\
& Alibaba/Qwen & -- & Qwen2.5-VL Series \\
& DeepSeek & -- & DeepSeek-VL2 \\
& Meta/Llama & -- & LLaMA4-Scout-17B \\
& Microsoft & -- & Phi-4-multimodal \\
\arrayrulecolor[gray]{0.7}\midrule\arrayrulecolor{black}
\multirow{4}{*}{\makecell[l]{\textbf{Open-source} \\ \textbf{LLMs}}}
& DeepSeek & DeepSeek-R1 & DeepSeek-V3 \\
& Alibaba/Qwen & Qwen3 Series & -- \\
& OpenAI & GPT-OSS-120B & -- \\
& Moonshot AI & Kimi-K2-Instruct & -- \\
\bottomrule
\end{tabular}
\end{table}

\subsection{Illustration of Medal Scorelines}
\label{appendix:medal_scoreline}

\begin{itemize}
    \item \textbf{IPhO, APhO, EuPhO:} These include separate theoretical and experimental exams, with medals awarded on total scores. As (M)LLMs cannot do experiments, we use the lowest theoretical exam score of gold medalists as the gold cutoff, and analogously for silver and bronze.  
    
    \item \textbf{NBPhO:} Theory and experiment appear in the same paper, with medals based on total scores. We set the gold line as the lowest total score of gold medalists. Since the paper contains experimental and plotting items, the model's full mark is lower than the human's; such items are counted as zero to reflect current limitations of (M)LLMs.  
    
    \item \textbf{PanPhO, PanMechanics:} These are theory-only exams, but the official website reports scores only for Hong Kong SAR contestants. Thus, thresholds are based on the lowest scores of Hong Kong SAR medalists. If scores of all medalists were available, the top-1 human score would likely be higher, while the medal thresholds could be lower.
    
    \item \textbf{F=MA:} With over 5,000 participants, only score histograms are published and no official medals are awarded. As a USAPhO qualifier, the cutoff score for advancement is treated as the gold line. Silver and bronze thresholds are inferred from Physics Bowl conventions (top 20\% and 35\%), estimated from the histogram, while the gold line also aligns with the Physics Bowl top-10\% rule.
\end{itemize}

\end{document}